%% file: main.tex
\documentclass[11pt]{article}
\usepackage[final]{acl}

\usepackage{times}
\usepackage{latexsym}
\usepackage[T1]{fontenc}
\usepackage[utf8]{inputenc}
\usepackage{microtype}
\usepackage{inconsolata}
\usepackage{graphicx}

\usepackage{amsmath,amssymb}
\usepackage{booktabs}
\usepackage{multirow}
\usepackage{xcolor}
\usepackage{enumitem}
\usepackage{morefloats}   
\usepackage{authblk}

\newcommand{\optcost}{\ensuremath{\mathcal{C}_{\text{opt}}}}

\title{Optimize Cheap, Deploy Strong: Cost-Aware Cross-Tier Transfer for Evolutionary Optimization}

\author{Tal Oved\thanks{Corresponding author: \texttt{Tal.Oved@ibm.com}.} \quad Roi Pony \quad Oshri Naparstek \quad Udi Barzelay \\
    IBM Research}

\begin{document}
\maketitle

\begin{abstract}
Evolutionary optimization of LLM prompts and agentic programs (e.g., GEPA) is
dominated by \emph{fitness evaluation}: scoring each candidate runs an answering LLM
over a validation set, so the evaluator's price tier dictates total
search cost. We restructure that search by \emph{decoupling the three roles} an LLM plays,
running the high-volume \emph{answering} role on the \emph{cheapest} tier, reserving
a \emph{strong} model for the rare \emph{reflection/variation operator}, then
exploiting \emph{upward cross-tier transfer} to deploy the cheaply evolved
prompt on a stronger target. We
contribute a cost-controlled characterization of \emph{when} cheap-tier search
substitutes for target-tier search, and where it fails. Across four tasks (HotpotQA,
IFBench, LiveBench-Math, HoVer) and eleven models in four model families, the
resulting prompt \emph{matches or exceeds} same-tier optimization while
placing over $\mathbf{96\%}$ of search tokens on the \emph{cheapest} tier, at
$\mathbf{5.6}$--$\mathbf{14\times}$ lower search cost, rising to
$\mathbf{25}$--$\mathbf{54\times}$ where reasoning tiers emit long chains of thought on
every fitness call.
\end{abstract}

\section{Introduction}
\label{sec:intro}
Evolutionary computation, population-based gradient-free search inspired by natural
selection, has a long track record on problems whose objective is black-box,
non-differentiable, or discrete, spanning genetic algorithms \citep{holland}, genetic
programming \citep{koza}, and neuroevolution \citep{neuroevolution}. Its defining loop
is the same everywhere: \emph{evaluate} a population's fitness, \emph{select}, and
\emph{vary} (mutate/recombine). A recurring theme across the field is that
\emph{fitness evaluation is the cost bottleneck}, which has motivated decades of work
on surrogate-assisted and multi-fidelity evolutionary algorithms that approximate the
expensive objective with a cheaper one \citep{chen2026efficient,labo}. Recently, this machinery has
been turned on a new discrete artifact: the natural-language prompts and programs that
drive large language models (LLMs).

In this LLM setting, automated prompt and program optimization has become a practical
alternative to manual prompt engineering, beginning with LLM-generated instruction search
\citep{zhou2023ape} and now covering declarative multi-stage programs whose instructions and
demonstrations are compiled by Bayesian search \citep{khattab2023dspy,opsahlong2024mipro} and
textual analogues of gradient descent \citep{yuksekgonul2024textgrad}. Evolutionary methods, which
maintain a population of candidate prompts, score them by a fitness function, and iteratively
mutate/select the best, are among the most effective, with reflective variants using
an LLM to read execution traces and propose targeted edits
\citep{gepa,evoprompt,promptbreeder,opro}.

Their central bottleneck is \emph{cost}, and it is widely reported as such: these methods
``require a substantial number of LLM calls and input tokens, making prompt optimization
expensive'' \citep{zehle2025capo}, and the same diagnosis of many round-trips and heavy token
volume motivates deliberately lightweight evolutionary search elsewhere
\citep{taherkhani2024epic}. The fitness of each candidate is estimated
by executing an \emph{answering} LLM over a validation set; to keep the estimate
statistically meaningful, the validation set cannot be small, so evaluation calls
overwhelmingly dominate the budget (Section~\ref{sec:problem}). Because those calls
run on the target model, optimizing a prompt for a strong/expensive model requires
running the \emph{entire} search on that model, capping population size, generations,
and ultimately quality for a fixed budget.

Here, we ask a simple question: \emph{must the search run on the model we intend to
deploy on?} We show the answer is no. Our method (Section~\ref{sec:method}) decouples the three
roles an LLM plays inside the evolutionary loop and assigns each its own tier: fitness is
evaluated by the \emph{cheapest} model available, the variation operator that proposes edits runs
on a \emph{strong} one, and the evolved prompt is deployed \emph{zero-shot} on a stronger target
tier, with no mapping, calibration or re-optimization. Upward cross-tier transfer has been observed
before for a single weak$\to$strong pair \citep{gepa,gao2026p1}; we study it systematically across
tiers, tasks and families, and turn it into a cost strategy.

\paragraph{Why the split is sound: an asymmetry of intelligence.} The two search roles make very
different demands. The variation operator must \emph{propose} high-precision edits, reading execution
traces and rewriting the prompt, which is the work that rewards a strong model, and it runs only
rarely. Fitness evaluation, run on every candidate, need only \emph{rank} prompts well enough to steer
selection rather than estimate any candidate's deployed quality precisely, which a cheap and noisy
evaluator does on the vast majority of comparisons. Spending strength where edits are proposed and
cheapness where fitness is merely ranked is therefore complementary rather than a compromise.

Two properties follow, and studying them together under controlled cost is our focus. First,
\textbf{cost reduction}: answering absorbs over $96\%$ of search tokens while the variation operator
is rare (Eq.~\eqref{eq:cost}), so moving \emph{only} the answering role to the cheapest tier moves
almost the whole bill. Existing cost reductions for LLM search cheapen the \emph{mutator}, cheapen the
\emph{evaluation of the same target}, or route models \emph{per query at inference time}
\citep{tanveer2026levi,zhao2025pmpo,zehle2025capo,chen2026efficient,labo,frugalgpt,routellm}; all of
them economize \emph{within} the tier the prompt will be served on, whereas we change \emph{which}
tier evaluates. Second, \textbf{zero-shot positive transfer}: one cheap search
deploys zero-shot across models of very different price and capability, which is notable given the
``model drift'' that degrades prompts moved laterally \citep{wang2025promptbridge}, and the cheap
prompt is often \emph{better} than the one the target tier optimized for itself, so the saving is not
bought with accuracy. Section~\ref{sec:analysis} locates that transferability in the variation operator
and reads the explicitness of the prompts it writes (Appendix~\ref{app:explicit}).


\vfill\eject 

\paragraph{Contributions.}
\begin{itemize}[leftmargin=1.6em,itemsep=1pt]
  \item A cost-aware reformulation that treats LLM tier as the fidelity dimension and decouples all
        three roles a model plays inside the loop, answering, variation, and deployment, studying them
        jointly as a cost/transfer trade-off (Section~\ref{sec:method}).
  \item Evidence over four tasks and eleven models in four model families that cheap search
        matches or beats same-tier optimization at $5.6$--$14\times$ lower cost on the Mixed Claude and GPT
        ladders and $25$--$54\times$ on Gemini (Section~\ref{sec:results}). At the limit a
        \emph{zero-API-cost} self-hosted Qwen3-8B answerer cuts the bill to the reflector's alone
        (under \$$2$ per run) and still matches each paid tier's own full-cost optimization.\footnote{%
        ``Zero-API-cost'' is exact and deliberately narrow: the model is open-weights and served
        locally, so its marginal price is \$$0$ in Table~\ref{tab:prices}. It is not free of compute.
        We served it through a shared internal vLLM~\citep{vllm} gateway whose GPUs we neither
        own nor meter, so
        every Qwen dollar figure is \emph{API} spend only; readers who would have to rent should add
        their own serving cost, modest for an 8B model at our call volumes but not quantified here.}
  \item A characterization of \emph{when} upward transfer succeeds: a measured explicitness gap in the
        prompts that transfer positively (Appendix~\ref{app:explicit}) and a role ablation locating the
        effect in the variation operator rather than the evaluator
        (Appendix~\ref{app:factorial}), together with the boundary conditions of near-zero cheap-tier
        competence below and the prompt-insensitive ceiling above, where prompt optimization of
        \emph{any} kind has no headroom \citep{gao2026p1}.
\end{itemize}

\section{Problem Statement and Cost Model}
\label{sec:problem}

\paragraph{Starting point: GEPA's optimization problem.} We build directly on the formulation of
\citet{gepa}. A compound AI system $\Phi$ pairs textual module prompts $\Pi$ (instructions $+$
demonstrations) with LLM weights $\Theta$, jointly its learnable parameters
$\langle\Pi,\Theta\rangle_\Phi$; a task instance $(x,m)$ induces an output scored by a metric $\mu$
valued in $[0,1]$, so over a task distribution $T$ the expected score is
\begin{equation*}
J_T(\Pi,\Theta)
  = \mathbb{E}_{(x,m)\sim T}\big[\mu\big(\Phi(x;\langle\Pi,\Theta\rangle_\Phi),\,m\big)\big].
\end{equation*}
Because a rollout (an invocation of $\Phi$ plus its $\mu$ evaluation) is expensive, the
sample-efficient form of the learning problem maximizes this score under a rollout budget $B$:
\begin{equation}
\label{eq:gepa}
\begin{aligned}
\langle\Pi^\ast,\Theta^\ast\rangle_\Phi
  &= \arg\max_{\langle\Pi,\Theta\rangle_\Phi}\, J_T(\Pi,\Theta)\\[2pt]
  &\quad\ \text{s.t. }\ \#\text{rollouts}\le B.
\end{aligned}
\end{equation}
This is Eq.~(2) of \citet{gepa}. GEPA keeps the weights \emph{frozen} and evolves only the prompts by
reflective mutation, and standard practice sets the frozen model to the deployment model,
$\theta_{\text{frozen}}=\theta_{\text{dep}}$, so the same LLM answers during search and serves at
deployment.

\paragraph{Our reformulation: split the frozen model into two roles, both $\neq M_{\text{dep}}$.}
The frozen weights enter Eq.~\eqref{eq:gepa}'s optimizer in \emph{two} distinct places: inside the
metric, where every fitness evaluation runs $\Phi$ on an \emph{answering} model; and inside the
reflective step, where a \emph{variation operator} model reads rollout traces and proposes prompt
edits. We assign these roles two separate frozen models, a cheap answerer $M_{\text{task}}$ and a
strong reflector $M_{\text{refl}}$, and let \emph{both} differ from the deployment model
$M_{\text{dep}}$. Search then maximizes a \emph{surrogate} objective
$J_{\text{task}}(\Pi)=J_T(\Pi,\theta_{\text{task}})$ while the quantity we care about is
$J_{\text{dep}}(\Pi)=J_T(\Pi,\theta_{\text{dep}})$, and writing the target as the surrogate plus a
residual makes the trade explicit:
\begin{equation}
\label{eq:decomp}
J_{\text{dep}}(\Pi)\;=\;\underbrace{J_{\text{task}}(\Pi)}_{\substack{\text{optimized cheaply,}\\\text{under }M_{\text{task}}}}
\;+\;\underbrace{\Delta(\Pi)}_{\substack{\text{cross-tier}\\\text{transfer residual}}},
\end{equation}
where $\Delta(\Pi):=J_{\text{dep}}(\Pi)-J_{\text{task}}(\Pi)$ is exactly the gap incurred by
searching with a cheaper model than we deploy on: we optimize against $J_{\text{task}}$ but are graded
on $J_{\text{dep}}$.

\paragraph{Two transfer quantities, kept distinct.} The residual $\Delta(\Pi)$ of
Eq.~\eqref{eq:decomp} is \emph{per-prompt}, for one \emph{fixed} prompt, and must not be confused with the \emph{cross-run} quantities we report.
Writing $\pi_s^\ast$ for the prompt optimized on a cheap source
tier $s$ and $\pi_t^\ast$ for the prompt optimized directly on the target tier $t$, define the
\emph{target regret} of cheap-tier search,
\[
R_{s\to t}=J_t(\pi_t^\ast)-J_t(\pi_s^\ast),
\]
where $s\to t$ denotes the \emph{direction of transfer} rather than an order of arguments. For
plotting we also name the residual in ``higher is better'' orientation and its scale-free form,
\begin{equation}
\label{eq:resid}
\delta_{s\to t}\;=\;-R_{s\to t},
\qquad
\delta^{\%}_{s\to t}\;=\;100\cdot\frac{\delta_{s\to t}}{J_t(\pi_t^\ast)},
\end{equation}
so $\delta^{\%}_{s\to t}>0$ means the cheaply-searched prompt \emph{beat} full same-tier optimization,
and normalizing by $J_t(\pi_t^\ast)$ removes differing score magnitudes across tasks and tiers. Our
core empirical claim is \emph{low $R_{s\to t}$ at substantially lower search cost}.

\paragraph{The two hypotheses we test.} Both are claims about the measured $R_{s\to t}$, which is
end-to-end: selection, the pool the mutator proposed and the deployment tier's own capability all
enter it.
\begin{itemize}[leftmargin=1.4em,itemsep=2pt,topsep=2pt]
\item \textbf{Zero-shot transfer (weak claim).} $R_{s\to t}\approx 0$: the cheaply-evolved prompt is
near-optimal on the deployment model, with no mapping, calibration or re-optimization.
\item \textbf{Zero-shot positive transfer (strong claim).} $R_{s\to t}<0$: it beats the prompt the
target tier optimized for itself, so transfer is not merely lossless but \emph{beneficial}. As in the
transfer-learning usage, ``positive'' is measured against the no-transfer baseline $\pi_t^\ast$, not
against the untrained seed, so it is a claim about the \emph{search}, not about the deployment tier's
raw capability.
\end{itemize}
Sections~\ref{sec:results}--\ref{sec:analysis} test both and find $R_{s\to t}$ small-to-favorable.
\emph{Why} positive transfer occurs is a separate question, taken up in Section~\ref{sec:analysis}.

\paragraph{Cost model: why a cheap $M_{\text{task}}$ is the lever.} A search attempts $A$ mutations,
each costing one token-heavy $M_{\text{refl}}$ call plus a cheap screening pass over a minibatch of
size $b$; the $K\le A$ attempts that survive screening are then scored on the full validation set of
size $N_{\text{val}}$, one $M_{\text{task}}$ call per (candidate, instance) pair. With $c(\cdot)$ the
\$/call cost,
\begin{equation}
\label{eq:cost}
\begin{aligned}
\optcost \;\approx\;\; &\underbrace{(K N_{\text{val}} + 2Ab)\, c(M_{\text{task}})}_{\text{fitness evaluation (dominant)}}\\[2pt]
        +\;\; &\underbrace{A \cdot c(M_{\text{refl}})}_{\text{variation (rare, token-heavy)}}.
\end{aligned}
\end{equation}
Because $N_{\text{val}}\gg b$ and only a fraction of attempts survive ($A/K\approx3$--$5$ in our runs),
the full-validation term dominates and the answering tier sets the budget. Breaking the coupling
$M_{\text{task}}=M_{\text{refl}}=M_{\text{dep}}$ drives $c(M_{\text{task}})$ to the \emph{minimum}
tier, cheapening that dominant term, while $M_{\text{refl}}$ stays strong for a \emph{bounded} premium
(it sits only in the second term, on a small call volume, though at a high enough per-token rate that
the premium is real; Section~\ref{sec:method}) and $M_{\text{dep}}$ is chosen freely at deploy time.
The savings are real only if the target regret they buy is small or negative, which is the empirical
question the rest of the paper settles.

\section{Method: Cost-Aware Cross-Tier Optimization}
\label{sec:method}
Our method is a drop-in modification to any reflective/evolutionary prompt optimizer
(we instantiate it on GEPA \citep{gepa}). It has three components.

\paragraph{(A) Cheap fitness tier.} Every candidate prompt is scored by running the
\emph{cheapest} answering model $M_{\text{task}}$ over the full validation set. The
cheap model is not a mathematical surrogate to be calibrated against the target; it
is the \emph{actual execution environment} during search. This makes the dominant
term of Eq.~\eqref{eq:cost} as small as possible.

\paragraph{(B) Strong variation operator.} New candidates are proposed by a
\emph{strong} reflector $M_{\text{refl}}$ that reads execution traces from the cheap
model and writes targeted edits: operator quality, not fitness-model quality, drives
improvement. What makes a strong model affordable is the volume asymmetry, not a high acceptance
rate (the operator fires on all $A$ \emph{attempted} mutations, $2.7$--$46\times$ more often than a
candidate is accepted): reflection is under $5\%$ of all search calls in every run
($0.08$--$4.95\%$). Those few calls still carry a median $27\%$ of a cheap run's spend (range
$12$--$46\%$ over $36$ runs), so the strong reflector is a bounded premium, not a free component.

\paragraph{(C) Upward cross-tier deployment.} The final evolved prompt is deployed
\emph{zero-shot} on the chosen target tier $M_{\text{dep}}$ (equal to or stronger
than $M_{\text{task}}$), with no learned mapping, calibration suite, or
re-optimization. One search thus amortizes across arbitrarily many deployment tiers.

\section{Experimental Setup}
\label{sec:setup}
\paragraph{Tasks.} Four tasks, each a faithful port of its GEPA-artifact setup, so that only the
models differ: \textbf{HotpotQA} (multi-hop QA; exact-match; DSPy $+$ BM25 retrieval, 2-hop program),
\textbf{IFBench} (instruction following; graded constraint-satisfaction; 2-stage program),
\textbf{LiveBench-Math} (competition math; deterministic LiveBench scorer; single chain-of-thought),
and \textbf{HoVer} (3-hop claim verification; binary recall of the gold supporting-document titles; a
4-call retrieval program reusing the BM25 index, with no answer LLM and no LLM judge). They span the
task types a cheap
answerer could plausibly fail on, so the thesis is tested rather than favorably sampled, and they span
the headroom available to prompt optimization at all, from a nearly saturated task to the benchmark
where it is most documented, which is what lets us separate our method failing from prompt
optimization having nothing to exploit (Appendix~\ref{app:detail}).

\begin{figure*}[t]\centering
\includegraphics[width=\linewidth]{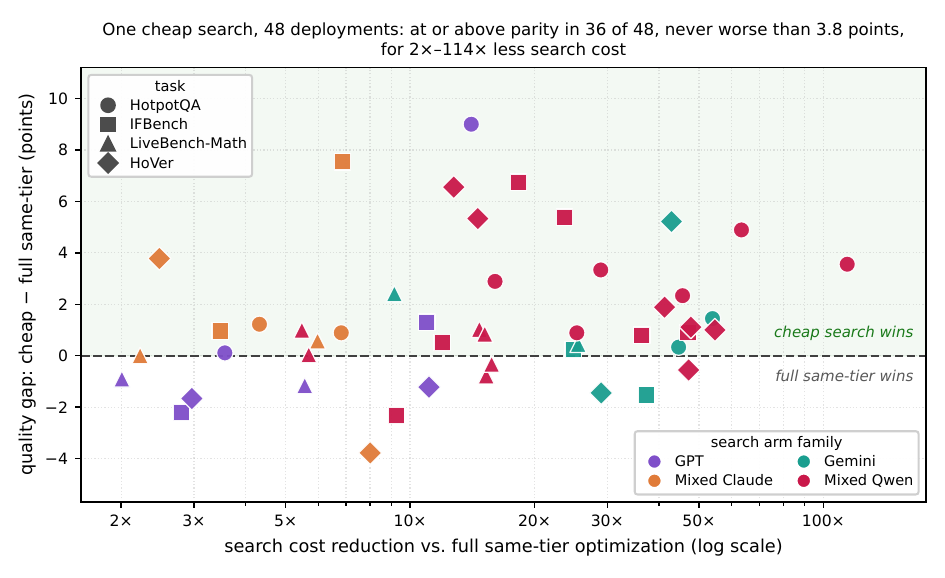}
\caption{Quality gap against full-cost optimization of the same tier ($y$; $0$ is parity) versus
search-cost reduction ($x$, log). One point per (task, search arm, deploy tier), $48$ in all. In the
shaded half-plane the cheap search won outright. Data in
Tables~\ref{tab:fam_claude}--\ref{tab:fam_qwen}.}
\label{fig:frontier}
\end{figure*}

\paragraph{Model tiers and methods compared.} \emph{Tier} means price rank
(Table~\ref{tab:prices}), not measured capability. Eleven models in four families, each named for the
model line that characterizes it, with the prefix \emph{Mixed} marking a family that draws on more
than one vendor. Three are paid ladders of rising price, each with its own strong
reflector: \textbf{Mixed Claude}
(gpt-4.1-nano $\to$ Claude Haiku 4.5 $\to$ Claude Sonnet-5; reflector Sonnet-5), an OpenAI answerer
under Claude reflection and Claude deployment, \textbf{GPT}
(gpt-4.1-nano $\to$ gpt-4.1-mini $\to$ gpt-5.6-luna; reflector gpt-5.5) and \textbf{Gemini}
(gemini-2.5-flash-lite $\to$ gemini-3.5-flash $\to$ gemini-2.5-pro; reflector gemini-3.1-pro). The
Mixed Claude and GPT ladders share the \emph{same} gpt-4.1-nano answerer, so gpt-4.1-mini also serves as a
neutral cross-family deploy target, while the Gemini ladder shares no model with either and is an
independent third-vendor replication. The fourth family is mixed in the same sense and is the
extreme case: \textbf{Mixed Qwen}, a self-hosted open-weights Qwen3-8B answerer under either paid
reflector (Sonnet-5 or gpt-5.5), deployed on the paid mini/Haiku/luna tiers, so the answering role
carries no per-token cost at all. Within each family we
compare \emph{Full-$X$}, full same-tier optimization ($M_{\text{task}}{=}M_{\text{refl}}{=}X$) and our
baseline, against \emph{Cheap+reflect$\to X$}, our method: the family's cheapest answerer plus its
strong reflector deployed zero-shot on $M_{\text{dep}}{=}X$. A \emph{seed-prompt control} (untrained
candidate~\#0, cost \$0) is reported on every deploy tier. All runs share identical data, program,
metric and budget per task, with caching disabled and $n{=}3$ seeds, and ``Deploy@$X$'' always
evaluates the validation-selected candidate, unchanged, on $X$'s held-out test set. Prices, per-task
budgets and the full protocol are in Appendix~\ref{app:impl}.

\section{Results}
\label{sec:results}
The per-benchmark tables, one per model family, and the cell-by-cell discussion behind them are
in Appendix~\ref{app:detail}. This section states what those tables establish and reads the three
figures that summarize them.

\begin{figure*}[t]\centering
\includegraphics[width=\linewidth]{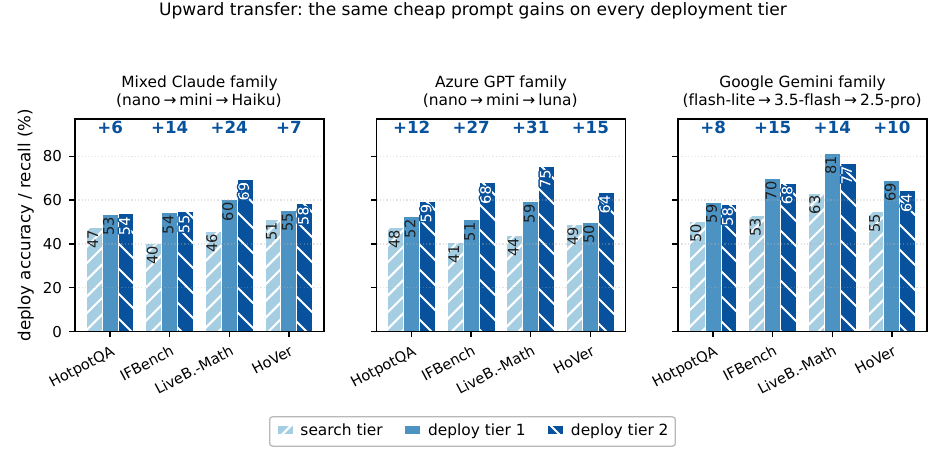}
\caption{Upward transfer of one cheap prompt, four tasks $\times$ three vendor families. Bars are the
search tier and the two deployment tiers, in the order named in each panel title; the bold label is the
net change. Tier order follows price, not measured quality: in the Gemini family the
newer \texttt{3.5-flash} outscores \texttt{2.5-pro} on all four tasks. Discussion in
Section~\ref{sec:results}.}
\label{fig:transfer}
\end{figure*}

\begin{figure}[t]\centering
\includegraphics[width=\columnwidth]{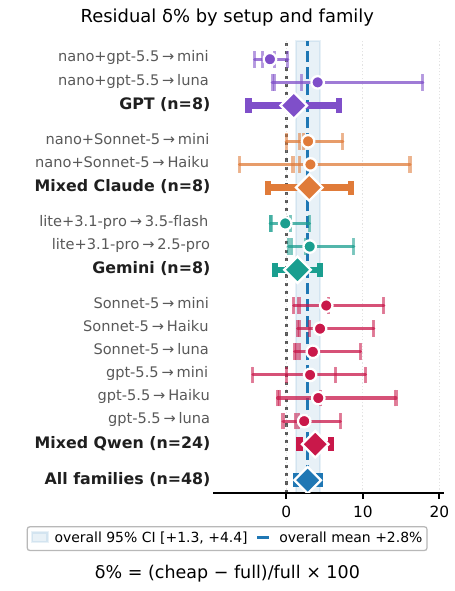}
\caption{Transfer residual $\delta^{\%}_{s\to t}$ (Eq.~\eqref{eq:resid}) over $48$ points, $12$ setups
$\times$ $4$ datasets; $\delta^{\%}_{s\to t}>0$ favours the cheap prompt. Circles are per-setup means,
ticks the individual datasets, thin bars their min--max range. Diamonds are per-family and pooled means
with $95\%$ $t$-intervals on those means. Statistics in Appendix~\ref{app:residual}.}
\label{fig:forest}
\end{figure}

\paragraph{The headline comparison.} A prompt searched with the cheapest answerer and a strong
reflector, then deployed one or two tiers up, matches or beats the prompt that same tier optimized
for itself at full price. This holds on four tasks, across eleven models in four families, and at
every deploy tier we tested: the cheap search is at or above parity in $36$ of the $48$ (task,
search arm, deploy tier) deployments, for $5.6$--$14\times$ less search cost on the Mixed Claude and GPT
ladders and $25$--$54\times$ less on Gemini, where both reasoning tiers emit long chains of thought
on every fitness call. The saving follows the volume: over $96\%$ of search tokens go to answering,
so moving that single role to the cheapest tier moves almost the whole bill while strength is still
bought where it is rare. At the limit, an open-weights answerer we host ourselves reduces the API
bill to the reflector's alone, under \$$2$ per run, and the resulting prompt still matches each paid
tier's own optimization. Figure~\ref{fig:forest} collects all $48$ into a single view, read in
Section~\ref{sec:analysis}.

\paragraph{The saving is not a tier discount.} Prices alone could explain the saving, so we measure how
far they would have to move to erase it. Let $\lambda^{\star}$ be the cheap-to-deployment price ratio at
which the cheap search stops being cheaper (Eq.~\ref{eq:lambda} in Appendix~\ref{app:impl}). It exceeds $1$ in $15$ of the $24$
cells with a full-cost own-tier search, so there the cheap composition stays cheaper even at price
parity: it emits fewer output \emph{tokens}, not merely cheaper ones. With quality already at parity it
is then the better buy. The full range is $0.50$--$3.46$, against $0.04$--$0.25$ in force
(Section~\ref{sec:limits}).

\paragraph{Quality against cost (Figure~\ref{fig:frontier}).} Reducing every deployment to a single
point makes the trade legible. The vertical axis is the quality gap against that tier's own
full-cost optimization, so parity is exactly the line $y{=}0$ and the half-plane above it is where
the cheap search wins outright; the horizontal axis is how much cheaper that search was. The points
sit on and above parity across two orders of magnitude of cost reduction, and in the minority of
cells that fall below it the shortfall is a few points at most. The gap also widens with the
strength of the deployment target, which is the direction the cost model of
Section~\ref{sec:problem} predicts and is visible as a slope rather than as an average.

\paragraph{Transferability (Figure~\ref{fig:transfer}).} The same evolved prompt, deployed unchanged
on its own search tier and then on both deployment tiers of its family, gains on all twelve task and
family pairs at zero extra search cost. Upward transfer is the reliable direction, unlike the
degradation reported for lateral transfer between comparable models~\cite{wang2025promptbridge}.
Price is not a proxy for quality: in $17$ of $75$ adjacent tier pairs the costlier model scores lower, so the costliest tier is not always the best target. And the
size of each jump is mostly the deployment model's own capability applied to the same prompt. The
figure shows that nothing breaks on the way up; the load-bearing quantity is the low target regret in
Appendix~\ref{app:detail}.

\begin{table}[t]\centering
\small\setlength{\tabcolsep}{4pt}
\begin{tabular}{lcc}
\toprule
 & median ratio & cheap higher \\[-1pt]
\midrule
Length (tokens) & $1.29\times$ & $40/52$ \\
Directives / 1k & $1.26\times$ & $35/52$ \\
Prohibitions / 1k & $1.17\times$ & $34/52$ \\
Capitalized words / 1k & $2.80\times$ & $36/48$ \\
\bottomrule
\end{tabular}
\caption{\textbf{The cheap search writes the more explicit prompt.} Median ratio of the cheap prompt to
the full-cost prompt it is paired with, over $52$ pairs. Lexicons and per-run counts in
Appendix~\ref{app:explicit}.}
\label{tab:explicit_main}
\end{table}

\paragraph{What the evolved prompts look like.} The cheap search writes the more explicit prompt.
Table~\ref{tab:explicit_main} pairs every cheap arm against the full-cost prompt of the tier it is
deployed on: the cheap prompt is longer and denser in directives, prohibitions and capitalized
emphasis, and all four markers move the same way. Section~\ref{sec:analysis} develops the account this
supports, that a weak evaluator rewards structure a stronger model then exploits.

\paragraph{Ablations and controls.} Several controls bound what the result rests on. A
zero-optimization row, the untrained seed prompt deployed on each tier at no cost, appears in every
table, and it is what separates tasks where prompt optimization has real headroom from tiers already
near a prompt-insensitive ceiling. A weak-reflector arm shows that cheapness alone is insufficient
and that operator strength is the lever. A $2\times2$ role ablation separates the two roles the
recipe changes together and locates the gain in the reflector (Section~\ref{sec:analysis},
Appendix~\ref{app:factorial}). A neutral deploy target that no reflector ever saw removes the overlap
between a family's costliest tier and its own reflector, and shows transfer is agnostic to the
reflector's family. Repeating the whole construction with MIPROv2 in place of GEPA shows the effect belongs to
the cost-aware setup rather than to one optimizer's mutation operator (Appendix~\ref{app:mipro}). We
also report the recurring side of the ledger, the query volume at which a longer prompt's serving
cost would offset the one-time search saving (Appendix~\ref{app:breakeven}), and the search curves
themselves (Appendix~\ref{app:curves}), which rank the cheap arms below the full-cost ones and so
make the case for judging these configurations on deployment quality rather than on search traces.

\begin{table}[!t]\centering
\small\setlength{\tabcolsep}{3pt}
\begin{tabular}{llccc}
\toprule
Evaluator & Reflector & @ Haiku & \$ & pts/\$ \\
\midrule
nano (cheap) & Haiku (weak) & $39.5\pm7.1$ & $1.95$ & n/a \\
\textbf{nano (cheap)} & \textbf{Sonnet-5} & $\underline{54.6\pm1.0}$ & $3.11$ & $\mathbf{13.0}$ \\
Haiku (full) & Haiku (weak) & $47.1\pm4.3$ & $22.17$ & $0.37$ \\
Haiku (full) & Sonnet-5 & $\mathbf{56.0\pm1.0}$ & $28.52$ & $0.62$ \\
\midrule
\multicolumn{2}{l}{\emph{reflector gain}, nano eval.} & $\mathbf{+15.1}$ & $+1.16$ & $13.0$ \\
\multicolumn{2}{l}{\emph{reflector gain}, Haiku eval.} & $\mathbf{+9.0}$ & $+6.35$ & $1.42$ \\
\bottomrule
\end{tabular}
\caption{\textbf{Role ablation on IFBench.} \% constraint-satisfaction at the Haiku deploy tier,
mean$\pm$std over $n{=}3$ seeds. The four rows differ only in the two model fields. pts/\$ is deploy
points per search dollar. Own-tier scores and cost basis in Appendix~\ref{app:factorial}.}
\label{tab:factorial_main}
\end{table}

\section{Analysis}
\label{sec:analysis}
Why does a prompt evolved on a \emph{weak} model transfer \emph{upward} cleanly, when lateral
transfer between models of comparable capability degrades \citep{wang2025promptbridge}?

\paragraph{The variation operator is where transferability is made.} Positive transfer is a property
of the role \emph{composition}: a strong variation operator makes the prompt transferable, and a cheap
evaluator makes producing it affordable. Table~\ref{tab:factorial_main} completes the $2\times2$ on
IFBench at identical budget, splits and token caps. The split is one-sided. Upgrading the
reflector is worth $+15.1$ points at the Haiku deploy tier, and it still pays when the evaluator is
already the full-cost one, so cheap evaluation is not what produces the gain. Restoring the full-cost
evaluator on top of a strong reflector buys little for a large bill, a $236\times$ difference in
return per search dollar. The recipe is therefore ``strong reflector, weak evaluator'' rather than
``use cheap models.'' Own-tier scores make this an account of \emph{transfer} rather than of prompt
quality: the stronger reflector's prompt is no better on the tier it optimized against, and much
better on a tier neither configuration searched against. The family tables bound the claim. A weak
reflector does not yield transferable prompts (Finding~1), and the reflector need not share a vendor
with the deployment model (Finding~5). Full table and cost normalization in
Appendix~\ref{app:factorial}.

\paragraph{What the transferable prompt looks like.} The composition says \emph{where} the effect is
produced. For \emph{why} a stronger tier exploits it we offer \textbf{structural explicitness} as a
hypothesis. A frontier model bridges logical gaps on its own, so a search run natively on it can
settle on an \emph{implicitly-coded} prompt that leans on the evaluator's reasoning. A cheap answerer
fails on ambiguity instead, so the reflector is pushed toward \emph{explicit and rigid} prompts:
guardrails, spelled-out constraints, edge cases. A frontier model absorbs that clarity easily. So
lateral transfer moves a prompt between differing latent biases, and upward transfer moves explicit
clarity into a model equipped to use it. The prompts do have that shape, on all four markers at once
(Table~\ref{tab:explicit_main}); both arms append validation answers in comparable amounts, so
memorization is not what separates them (Appendix~\ref{app:prompts}). Two limits apply. This is a
correlation, not an ablation. And the pairing differs in \emph{both} roles, so it cannot separate the
weak evaluator forcing the explicitness from the strong reflector writing it, though the factorial
favours the reflector. Appendix~\ref{app:discussion} lists what would settle it.

\paragraph{The residual concentrates on zero.} Figure~\ref{fig:forest} aggregates the method into one
picture: every cross-tier deployment we ran, pooled as paired residuals $\delta_{s\to t}=-R_{s\to t}$
of Eq.~\eqref{eq:resid} in the scale-free form $\delta^{\%}_{s\to t}$, so tasks with different score
ranges are commensurable. The mean residual is ${+}2.8\%$ of the full-cost score in favour of the
cheap search, with a $95\%$ interval above zero. Searching cheaply and deploying upward is to first
order outcome-equivalent to paying for full same-tier optimization, at roughly an order of magnitude
less cost. Every family's own mean is positive and all four intervals overlap, so this is a property
of the construction rather than of one favourable ladder. The shape matters more than the mean. The
downside is bounded and shallow: no setup collapses, and every point is covered once the allowed
shortfall reaches about six percent of the full-cost score (Figure~\ref{fig:coverage},
Appendix~\ref{app:residual}). The upside carries a genuine right tail, so where prompt optimization
still has headroom the cheap prompt wins outright. That asymmetry is what the explicitness account
implies: spelled-out instruction is neutral where a tier is already saturated and a multiplier where a
strong deployment model has room to exploit it.

\section{Related Work}
\label{sec:related}
Our work sits where three lines meet: discrete evolutionary search, multi-fidelity efficiency, and
cross-model prompt transfer.

\paragraph{Evolutionary Prompt Optimization.} PromptBreeder \citep{promptbreeder} and EvoPrompt
\citep{evoprompt} established LLMs as variation operators over the discrete space of natural-language
prompts; GEPA \citep{gepa}, which we build on, adds reflective trace-driven mutation, Mage
\citep{mage} extends it with multi-objective Pareto selection, and OPRO \citep{opro} exploits the
optimization trajectory as meta-data. GEPA already exposes a separate reflection model, so
answerer/reflector decoupling is not by itself new. What stays open is that the answerer, reflector
and deployment model are still instantiated at the \emph{same} tier, so every fitness
evaluation scales with the target's inference cost, and no cost-controlled account establishes
\emph{when} cheap-tier search substitutes for target-tier search. We treat the three as separately
chosen axes and characterize that regime.

\paragraph{Multi-Fidelity and Computational Efficiency.} Cascades that route queries at test time cut
inference cost \citep{frugalgpt,routellm}, successive halving splits a budget across configurations
\citep{hyperband}, and surrogate-assisted evolutionary computation approximates an expensive objective
with a cheaper one \citep{surrogate,chen2026efficient,labo}. Fidelity there is data-subset size,
budget, or a surrogate of the \emph{same} objective. Cost-aware LLM search does likewise: PMPO
cheapens scoring while still running on the \emph{target} \citep{zhao2025pmpo}, and CAPO and EPiC hold
down candidate and call counts \citep{zehle2025capo,taherkhani2024epic}. A second group instead varies
\emph{which} model acts inside the search, by mutation role \citep{tanveer2026levi}, by a bandit over
generators \citep{lange2026shinka}, by generation confidence \citep{ray2026adaptevolve}, or by relaying
a cheap population to a strong model mid-run \citep{luo2026relay}. Each picks a model for a step, so
the strong tier keeps spending search budget. We make fidelity the \emph{model tier}
itself: the cheap model is the actual execution environment during search, strength is reserved for
the rare variation operator, and the deployment model changes with no calibration to the target.

\paragraph{Cross-Model Transferability.} Prompt portability is limited by ``model drift''
\citep{wang2025promptbridge}: lateral transfer between models of comparable capability degrades
sharply, and while soft-prompt transfer addresses embedding shifts \citep{wang2025efficient}, discrete
prompts are usually assumed model-specific. Weak$\to$strong evidence does exist, in GEPA-Qwen-Opt
\citep{gepa} and in \citet{gao2026p1}, who optimize on Qwen3-4B and deploy on Qwen3-30B, but in both
it is an incidental generalization
observation on a single model pair rather than a cost strategy. We instead test upward portability
systematically and turn the property into an order-of-magnitude cost reduction; we are not aware of
prior work combining cheap evaluation, strong mutation and upward deployment as a cost strategy and
characterizing when it holds.

\section{Conclusion}
\label{sec:conclusion}
Decoupling the fitness-evaluation tier from the variation operator and from the deployment model turns
the dominant cost of evolutionary prompt search into a free parameter. Across four tasks and eleven
models in four families, a cheap search with a strong reflector, deployed upward, matches or beats
same-tier optimization at $5.6$--$14\times$ lower cost, spending over $96\%$ of its search tokens on
the cheapest tier, and yields a tier-portable ``compile-once, deploy-anywhere'' prompt, precisely on
the tasks and tiers where prompt optimization has headroom to begin with. Weak$\to$strong prompt
transfer has been observed before \citep{gepa,gao2026p1}; our contribution is to characterize
systematically \emph{when} cheap-tier search substitutes for target-tier search, to show that the
substitution is \emph{positive} rather than merely lossless, and to locate its source in the variation
operator rather than in cheap evaluation.

\section{Limitations}
\label{sec:limits}
The method is bounded by the cheap tier's baseline competence. If the cheap model
scores $\approx 0$ on the task, the fitness landscape is flat (no gradient, no
tracebacks, no selection pressure) and search stagnates. The cheap evaluator must
achieve at least marginal success given a good prompt.

\paragraph{Low-headroom (prompt-insensitive) deployment tiers.} The opposite boundary is a target
tier already near its prompt-insensitive ceiling. On LiveBench-Math the untrained seed prompt scores
$41.5/58.8/72.1$ (nano/mini/luna), within $1$--$4$ points of every optimized method, and validation
curves are correspondingly flat (Finding~4). This is a limitation of \emph{evolutionary prompt
optimization in general} rather than of cost-aware transfer: \citet{gao2026p1} likewise report
model/task combinations where an optimized prompt does not beat the untuned base prompt at all.

\paragraph{Dependence on the price schedule.} Every dollar figure is \emph{derived}: we measure
per-call token counts per model and apply Table~\ref{tab:prices} as a lookup, so any future schedule
can be re-applied without re-running anything. The honest question is not whether the multiplier
stays at $5.6$--$14\times$ but how far prices would have to move to overturn it, which is what the
break-even ratio $\lambda^{\star}$ of Section~\ref{sec:results} measures: the tier gap has to vanish
entirely, and in $15$ of $24$ cells even that is not enough. The conclusion is thus contingent on
providers offering a tiered lineup at all, not on any specific rate, and it reproduces under three
independently set schedules (Anthropic, Azure, Google).

\clearpage


\bibliography{references}


\clearpage

\appendix
\raggedbottom
\section{Implementation Details}
\label{app:impl}

\begin{table}[h!]
\centering
\footnotesize\setlength{\tabcolsep}{4pt}
\begin{tabular}{@{}llcc@{}}
\toprule
Model & Role & in & out \\
\midrule
gpt-4.1-nano     & answerer (Mixed Cl., GPT) & $0.10$ & $0.40$ \\
gpt-4.1-mini     & GPT mid / neutral target & $0.40$ & $1.60$ \\
Claude Haiku 4.5 & Mixed Cl.\ mid & $0.76$ & $3.80$ \\
Claude Sonnet-5  & Mixed Cl.\ refl.\ + mid & $1.52$ & $7.60$ \\
gpt-5.5          & GPT reflector & $2.5$  & $15$   \\
gpt-5.6-luna     & GPT mid & $1$    & $6$    \\
\midrule
gemini-2.5-flash-lite & Gem.\ answerer & $0.10$ & $0.40$ \\
gemini-3.5-flash & Gemini mid & $1.50$ & $9.00$ \\
gemini-2.5-pro   & Gemini mid & $1.25$ & $10.0$ \\
gemini-3.1-pro   & Gemini reflector & $2.00$ & $12.0$ \\
\midrule
Qwen3-8B         & local answerer, unpriced & $0$    & $0$    \\
\bottomrule
\end{tabular}
\caption{Per-model token prices (USD per 1M tokens, \emph{in}put/\emph{out}put) and roles, as published
at the time of the experiments. Every cost in the paper is computed from these with caching off.
Qwen3-8B is served locally, so its marginal token price is \$$0$ and that family's whole bill is the
reflector's; the serving compute is excluded throughout. No deployment target is priced above its own
family's reflector, which is why we call every deploy tier a \emph{mid} tier; the \emph{out} column,
not the role label, carries the ordering.}
\label{tab:prices}
\end{table}

\paragraph{Protocol.} Identical data (train/val/test), program, metric, and budget per task; only the
models differ. Optimizer = GEPA at paper scale, with per-task budgets of HotpotQA $6871$, IFBench
$3593$, LiveBench-Math $1839$, and HoVer $7051$ metric calls. $n{=}3$ seeds: HotpotQA's split is
seed-dependent (a fresh train/val/test draw per seed), whereas the IFBench, LiveBench-Math, and HoVer
splits are fixed, so there the seed varies only GEPA's search RNG. Optimization cost is exact
(per-call token logging $\times$ published prices, cache off, a true cost). Costs are therefore
reproducible under price revision: the logged quantity is \emph{tokens per call per model}, and any
future price schedule can be applied to every run in this paper without re-execution. 

\paragraph{Break-even price ratio.} Let $\lambda$ scale the cheap answering tier's input and output rates
together, as a fraction of the deployment tier's, so $\lambda{=}1$ is price parity; the ratio in force in
our experiments is $0.04$--$0.25$. Holding the measured token counts fixed, the cheap arm costs
$\lambda A_{\text{cheap}} + R_{\text{cheap}}$, where $A_{\text{cheap}}$ is its answering tokens priced at
the deployment tier's own rate and $R_{\text{cheap}}$ its reflector spend, which sits on a third tier and
does not move with $\lambda$. The full-cost arm costs $C_{\text{full}}$, independent of $\lambda$.
Equating the two gives the break-even ratio
\begin{equation}
  \lambda^{\star} = \frac{C_{\text{full}} - R_{\text{cheap}}}{A_{\text{cheap}}},
  \label{eq:lambda}
\end{equation}
below which the cheap search is the cheaper one. Over the $24$ cells that have a full-cost own-tier
search $\lambda^{\star}$ runs $0.50$--$3.46$, so the cheap tier's price would have to rise by
$2.7$--$86\times$ before a cell turns. In $15$ of the $24$ it exceeds $1$: there the cheap arm also emits
fewer output tokens over the whole search than the full arm does, so it stays cheaper at parity and the
saving is a property of the composition rather than of the price list. In the remaining nine the token
volumes are close or reversed, and the saving does rest on some discount. Mixed Claude on HotpotQA
deployed at Haiku is a worked case: $C_{\text{full}}{=}\$40.77$, $R_{\text{cheap}}{=}\$1.97$ and
$A_{\text{cheap}}{=}\$33.00$ give $\lambda^{\star}{=}1.18$, against $0.11$ in force. Two caveats:
$\lambda$ is a single scalar on both rates, so $\lambda^{\star}$ idealizes a two-rate price list, and it
speaks only to cost. Quality enters separately, through $\delta_{s\to t}$.
``Deploy@$X$'' takes
the candidate selected by its \emph{validation} score during search and evaluates it, unchanged,
on $X$ over the held-out \emph{test} set; selection uses validation only, so there is no
test-set selection.

\paragraph{Uniform measurement conditions.} Within every comparison, both prompts are scored on
the same held-out test set, through the same harness, at the same output-token cap: $3000$ for
HotpotQA, $4000$ for IFBench, $16384$ for LiveBench-Math and HoVer. Caps differ across benchmarks and
we never compare token counts across benchmarks. The prompt evaluated in each cell is the run's best
candidate by mean per-instance validation subscore, the rule used everywhere in this paper. A cap that binds asymmetrically would put the
output-token deltas of Table~\ref{tab:breakeven} in question, so we measured how often one binds: in
eleven of its twelve rows the at-cap rate is at most $8.1\%$ and near-symmetric across the two arms.
IFBench at $3.5$-flash binds on $38.3\%$ and $37.1\%$ of calls, and recomputing that row's delta over
the uncensored calls alone reproduces the full-sample value, so the cap is not what drives it.

\paragraph{Deployment parse handling (adaptive-thinking models).} One deployment target,
gpt-5.6-luna, is an adaptive-thinking model that intermittently omits the chain-of-thought
\texttt{reasoning} field DSPy's structured-output adapter expects, so the adapter rejects a
response whose answer field is present. This is a decoding artifact of that one model, absent for
every other, and not a failure to solve the task. We therefore recover the intended field from the
raw completion, so the parse does what it should. It succeeded on every affected step, so no
instance is dropped or scored zero and the luna deploy runs the full multi-hop pipeline on the same
$300$-example test set as its baselines. The comparison stays fair and the numbers are clean.

\section{Optimizer Ablation: MIPROv2}
\label{app:mipro}

\begin{table*}[t]\centering
\small\setlength{\tabcolsep}{3.2pt}
\begin{tabular}{llccccccc}
\toprule
\multicolumn{2}{c}{Method} & \multirow{2}{*}{opt cost \$} & \multirow{2}{*}{cost$\downarrow\times$} & \multirow{2}{*}{test @ nano} & \multirow{2}{*}{test @ mini} & \multirow{2}{*}{test @ Haiku} & \multirow{2}{*}{test @ luna} & \multirow{2}{*}{test @ Sonnet} \\
\cmidrule(lr){1-2}
Task model & Proposer model &  &  &  &  &  &  & \\
\midrule
\multicolumn{9}{l}{\emph{HotpotQA}} \\
\cmidrule(l){1-9}
gpt-5.6-luna & gpt-5.6-luna & $77.1$ & $1.0\times$ & -- & -- & -- & $\mathbf{57.1\pm1.3}$ & -- \\
Haiku & Haiku & $79.8$ & $1.0\times$ & -- & -- & $53.3\pm3.9$ & -- & $52.7\pm4.3$ \\
gpt-4.1-mini & gpt-4.1-mini & $29.6$ & $2.7\times$ & -- & $51.3\pm2.3$ & $51.4\pm2.8$ & $54.2\pm1.8$ & -- \\
\textbf{nano} & \textbf{gpt-5.5} & $\mathbf{8.5}$ & $\mathbf{9.4\times}$ & $47.4\pm1.2$ & $51.2\pm6.7$ & $51.9\pm4.0$ & $52.7\pm6.8$ & -- \\
\textbf{nano} & \textbf{Sonnet-5} & $\mathbf{9.3}$ & $\mathbf{8.5\times}$ & $47.4\pm2.4$ & $\mathbf{52.8\pm6.2}$ & $\mathbf{54.3\pm6.4}$ & $\mathbf{57.1\pm1.2}$ & $\mathbf{60.9\pm1.8}$ \\
\midrule
\multicolumn{9}{l}{\emph{IFBench}} \\
\cmidrule(l){1-9}
gpt-5.6-luna & gpt-5.6-luna & $40.7$ & $1.0\times$ & -- & -- & -- & $66.1\pm1.1$ & -- \\
Haiku & Haiku & $31.1$ & $1.3\times$ & -- & -- & $52.4\pm4.6$ & -- & $72.4\pm2.1$ \\
gpt-4.1-mini & gpt-4.1-mini & $10.6$ & $3.8\times$ & -- & $50.5\pm1.3$ & $52.6\pm1.4$ & $67.1\pm0.7$ & -- \\
\textbf{nano} & \textbf{gpt-5.5} & $\mathbf{2.9}$ & $\mathbf{14.2\times}$ & $42.0\pm0.4$ & $51.7\pm3.5$ & $55.7\pm1.9$ & $\mathbf{70.2\pm1.0}$ & -- \\
\textbf{nano} & \textbf{Sonnet-5} & $\mathbf{3.3}$ & $\mathbf{12.2\times}$ & $44.3\pm0.7$ & $\mathbf{55.0\pm1.6}$ & $\mathbf{57.5\pm3.0}$ & $\mathbf{70.2\pm1.0}$ & $\mathbf{75.0\pm1.9}$ \\
\bottomrule
\end{tabular}
\caption{\textbf{Optimizer ablation: the same decoupling run with MIPROv2 instead of GEPA.} The two
cheap rows (bold) are the body's construction with \texttt{dspy.MIPROv2} in place of GEPA's reflective
mutation. Conventions as in Table~\ref{tab:fam_gpt}; ``--'' $=$ not run on that tier. Configuration and
two caveats in the accompanying text.}
\label{tab:mipro}
\end{table*}

Every result in the body uses GEPA as the evolutionary optimizer, so a natural question is whether the
cost-aware decoupling is a property of the \emph{setup} (cheap answerer, strong proposer, cross-tier
deployment) or an artifact of GEPA's reflective-mutation operator. Table~\ref{tab:mipro} repeats the
construction with a structurally different optimizer: \texttt{dspy.MIPROv2}
\citep{opsahlong2024mipro}, which performs Bayesian
search over instructions \emph{and} few-shot demonstrations rather than LLM-authored mutations. The role
that GEPA fills with a reflection model, MIPROv2 fills with an instruction \emph{proposer}, so the cheap
rows keep the same shape as in the body: a \texttt{gpt-4.1-nano} answerer scored over the validation
set, with a strong model called only to propose, and the resulting prompt deployed unchanged on
stronger tiers.

\paragraph{Configuration.} MIPROv2 runs at \texttt{auto=heavy} with $4$ bootstrapped and $4$ labeled
demonstrations, \texttt{minibatch\_size}$=35$ and caching off, on the same data, program and metric as
the body's runs; only the optimizer changes. Deploy token caps match the body's fair-rescore caps:
reasoning targets (luna, Sonnet) at $12000$ tokens and non-reasoning targets (mini, Haiku) at $4000$,
with MIPROv2's own-tier luna runs already using the $12000$ cap in both roles. The columns span both
vendors' ladders because the MIPROv2 deploy matrix does.

\paragraph{What the swap shows.} The two cheap configurations are by far the cheapest rows in both
benchmark sections, at \$$2.9$--\$$9.3$ against \$$10.6$--\$$79.8$ for full same-tier search, and they
still take the best mean in seven of the eight deploy columns and draw level in the eighth.
Cheap-tier search substituting for target-tier search therefore does not depend on GEPA's mutation
operator.

\paragraph{Two caveats on reading the table.} First, search cost is \emph{not} budget-matched to the
body's GEPA runs. MIPROv2's rollout count is emergent from \texttt{auto=heavy} rather than capped at a
metric-call budget, so the \$ column is comparable \emph{within} this table, which is what
cost$\downarrow\times$ measures, but not across optimizers. This is an ablation of the
\emph{mechanism}, not an optimizer race, and a per-configuration MIPROv2 versus GEPA head-to-head is
outside its scope. Second, not every lead in the table is resolved: the HotpotQA cheap rows carry a
$\pm6$--$7$ point seed spread from \texttt{nano}'s own variance, so their leads there sit within noise,
whereas the IFBench leads ($1.4$--$3.1$ std) do not.

\section{Role Ablation: Which Role Carries the Gain?}
\label{app:factorial}

\begin{table*}[t]\centering
\small\setlength{\tabcolsep}{4pt}
\begin{tabular}{llccccl}
\toprule
Evaluator & Reflector & test @ nano & test @ Haiku & search \$ & pts/\$ & \\
\midrule
nano (cheap) & Haiku (weak) & $41.2\pm2.6$ & $39.5\pm7.1$ & $1.95\pm0.13$ & n/a & \emph{reference} \\
\textbf{nano (cheap)} & \textbf{Sonnet-5} & $40.2\pm1.3$ & $\underline{54.6\pm1.0}$ & $3.11\pm0.17$ & $\mathbf{13.0}$ & \emph{our configuration} \\
Haiku (full) & Haiku (weak) & -- & $47.1\pm4.3$ & $22.17\pm1.48$ & $0.37$ & \\
Haiku (full) & Sonnet-5 & -- & $\mathbf{56.0\pm1.0}$ & $28.52\pm1.51$ & $0.62$ & \emph{best, $9.2\times$ the \$} \\
\midrule
\multicolumn{2}{l}{\emph{reflector contrast} @ nano eval.} & $-1.0$ & $\mathbf{+15.1}$ & $+1.16$ & $13.0$ & \\
\multicolumn{2}{l}{\emph{reflector contrast} @ Haiku eval.} & -- & $\mathbf{+9.0}$ & $+6.35$ & $1.42$ & \\
\multicolumn{2}{l}{\emph{evaluator contrast} @ Sonnet-5 refl.} & -- & $-1.4$ & $+25.41$ & $0.055$ & \\
\bottomrule
\end{tabular}
\caption{\textbf{Role ablation: evaluator $\times$ reflector on IFBench, full version.} \%
constraint-satisfaction, mean$\pm$sample std over $n{=}3$ seeds; the four configurations differ only in
the two model fields. \emph{test @ nano} is the own-tier score, blank where the evaluator's own tier is
the deploy tier. Search \$ is per run, normalized for token-log coverage (see \emph{Cost basis}).
pts/\$ is deploy points per search dollar, against row 1 for the configurations and against the paired
cell for the contrasts.}
\label{tab:factorial}
\end{table*}

Our cheap configuration changes \emph{two} roles at once relative to full same-tier search: the
evaluator drops to \texttt{gpt-4.1-nano} and the reflector rises to \texttt{claude-sonnet-5}. Since
both move together, no single comparison in the family tables can say which one carries the result.
Table~\ref{tab:factorial} completes the $2\times2$ on IFBench with the two missing cells;
Table~\ref{tab:factorial_main} in \S\ref{sec:analysis} is the same experiment abridged, and this
appendix adds the own-tier column, the cost normalization and the caveats. All four configurations are
identical in budget ($3593$ metric calls), splits and token caps; only the two model fields differ.

\paragraph{The reflector is the lever, and it buys \emph{transfer}.} Upgrading only the reflector is
worth $+15.1$ points at the Haiku deploy tier with a cheap evaluator and $+9.0$ with a full one, so
the gain is not an interaction peculiar to cheap evaluation. The two columns say what kind of gain it
is. On the tier it optimized on (\emph{test @ nano}) the weak-reflector run is \emph{not} worse: both
prompts are equally good at the objective they were selected against. Deployed unchanged on Haiku, a
tier neither optimized on, they separate by $15$ points, and the variance separates with them. So the
strong reflector does not fit the search environment better; it writes a prompt that survives the move
upward. The weak-reflector prompt is specialized to its cheap evaluator, invisibly on-tier and
expensively off-tier. That is the mechanism of \S\ref{sec:analysis} seen by ablation rather than
correlation.

\paragraph{A strong reflector alone is not the recommendation.} The best cell in the square is the
expensive one, Haiku evaluator $+$ Sonnet-5 reflector, but only just, and the \$ column prices that
margin: the reflector upgrade returns $13$ points per dollar, additionally restoring the full-cost
evaluator returns $0.055$, a $\times236$ difference. The reason is volume. The evaluator runs on every
validation instance of every generation while the reflector fires a few dozen times, so it is $>96\%$
of the tokens (\S\ref{sec:results}). Strength is worth paying for in the reflector and close to
worthless in the evaluator. Table~\ref{tab:factorial_main} states that recipe in the body; this
ablation on GEPA is what establishes it.

\paragraph{Cost basis.} The \$ column is normalized, because the four arms were not logged equally
well and a partial cost log biases a ratio rather than just adding noise to it. The two pre-existing
arms predate a fix to our token accounting and record $74\%$ of their per-call tokens (a completion
served from cache skips the logging callback), against ${>}107\%$ for the two new arms, so we divide
each arm's raw total by its own measured coverage. Normalizing \emph{shrinks} the expensive/cheap ratio
from $12.9\times$ to $9.2\times$, the conservative direction, and puts full-Haiku search at $7.1\times$
the cheap arm, reproducing the independently measured ${\sim}7\times$ of Table~\ref{tab:fam_claude}.
Raw totals are \$16.39, \$2.39, \$2.10 and \$30.74 in row order.

\section{Deployment Break-Even Volume}
\label{app:breakeven}
Our savings are a \emph{one-time}
reduction in \emph{search} cost, whereas a longer evolved prompt costs more on \emph{every} deployed
query, so at high enough volume the recurring cost could erode, or erase, the one-time saving. We
measure both sides. From the per-call token logs of the same held-out scorings, we compute the
per-query token delta between the cheap-search prompt and the full-cost prompt on the same deploy
model, and solve for the break-even volume
\begin{equation}
N^\star = \frac{\Delta C_{\text{search}}}{c_{\text{query}}^{\text{cheap}} -
c_{\text{query}}^{\text{full}}},
\label{eq:breakeven}
\end{equation}
the query count at which the two total-cost curves cross.
Table~\ref{tab:breakeven} reports, for each (benchmark, deploy tier) cell, the one-time search saving
$\Delta C_{\text{search}}$, the per-query input/output token deltas of the cheap-search prompt
relative to the full-cost prompt on the \emph{same} deploy model, and the resulting $N^\star$ of
Eq.~\eqref{eq:breakeven}.

\paragraph{What each cell compares.} Every row is an \emph{own-tier} comparison at $n{=}3$ seeds: the
cheap-search prompt against the full-cost prompt evolved on the very model both are then deployed on.
This is the comparison the claim is about (``was the cheap search worth it, versus paying full price
on this tier?''), so we run it uniformly rather than substituting a cross-tier prompt where own-tier
data was inconvenient to obtain. Two consequences are worth stating. First, the table covers exactly
the tiers on which a full-cost search was actually run: Haiku and luna in the Mixed Claude and GPT families,
and 3.5-flash in the Gemini one. Second, the cheap arm of each row is the family's cheap search for that deploy
model (nano${+}$Sonnet-5 for the Haiku rows, nano${+}$gpt-5.5 for the luna rows, and
flash-lite${+}$3.1-pro for the Gemini one), which is the arm reported for that tier in the family
tables of Appendix~\ref{app:detail}, not a per-cell pick of whichever cheap run happened to look best.
The Gemini rows are the method's \emph{hard} case rather than a favourable one: that family's cheap
search is so much cheaper than its full-cost search that $\Delta C_{\text{search}}$ is an order of
magnitude larger than in the other families, and a larger one-time saving takes \emph{more} deployed
queries to pay back, pushing $N^\star$ up rather than down.

\paragraph{Token-log completeness.} A per-query token mean is only meaningful if the log covers the
whole scoring, so we require a complete per-call log for all three seeds on both arms of every cell,
and we re-scored the cells that missed that bar. A partial log biases the mean rather than thinning it,
because a module's stages carry different prompt lengths, and that shifts tokens per call, the
denominator of $N^\star$. Counting calls is not enough, since a scoring that abandoned examples on a
timeout still reaches roughly the expected file size, so we also require that every held-out example
scored. The failures were confined to the two longest per-example chains, HoVer and HotpotQA, and we
re-scored those cells at a longer per-example limit. The measured bias is small. Re-scorings replay the
same candidate prompts under the same caps, so they change no accuracy number in this paper, only
token counts.

\paragraph{Effect of prompt caching.} Under prompt caching, which is appropriate here since the evolved
instruction is a stable prefix, the input
delta is billed at $0.1\times$, and the direction this moves the crossing is \emph{not} uniform, because
the discount applies to whichever arm spends more input. Where the cheap prompt is the longer one
($\Delta\text{in}>0$) its per-query penalty shrinks and the crossing recedes, in one cell by more than
five-fold (HotpotQA@Haiku, $50$k to $277$k); where the cheap prompt \emph{saves} input
($\Delta\text{in}<0$) that saving shrinks too and the crossing arrives sooner, in one cell by a factor of
nearly three (LiveBench-Math@3.5-flash, $137$k to $48$k). Caching is therefore not a free margin of
safety for our claim in every cell, and we report both columns rather than the more favourable one. The
honest summary is that our claim is a search-cost claim with a volume caveat that is material only for
high-volume deployments of a task whose full-cost search was already inexpensive.
Relatedly, Eq.~\eqref{eq:cost} separates the (dominant)
fitness-evaluation term from the (rare) variation term, and Appendix~\ref{app:detail} reports the measured
per-run split: the reflector is $0.3$--$1.2\%$ of calls and $1.5$--$4\%$ of tokens but
$28$--$35\%$ of spend.

\paragraph{Search-cost coverage.} $\Delta C_{\text{search}}$ is a mean over the seeds per arm, taken
from the search-side per-call logs. 

\paragraph{Three details of the measurement.} First, per-query cost is taken from the real per-call token
logs of the held-out scorings (one $\{$input, output, model$\}$ record per LM call), not from a
character heuristic on the instruction string: the billed prompt is the framework-rendered one
(signature, field templates, retrieved context), of which the evolved instruction is only a part.
Second, we use \emph{mean tokens per call} multiplied by the module's architectural calls per query
($5$ for HotpotQA, $4$ for HoVer, $2$ for IFBench, $1$ for LiveBench-Math), so that a transient retry
, which duplicates calls but not prompt size, cannot inflate the estimate; recomputing from the
raw per-example totals instead moves individual cells but changes no sign. Third, the analysis needs
per-call token logs on \emph{both} arms of a cell, and the Qwen deploy driver did not write them, so
that family is absent here rather than approximated; the tiers reported are the Mixed Claude, GPT and Gemini
ones on which a full-cost search was run, at all four benchmarks each.

\input{tab_breakeven}

\clearpage

\section{Explicitness Profile of the Evolved Prompts}
\label{app:explicit}
\label{app:lexicons}
Table~\ref{tab:explicit} counts explicitness markers in each run's best candidate, chosen by the same
argmax-mean-validation-subscore rule the deployment scripts use, concatenated over the module's
predictors and tokenized with the \texttt{o200k\_base} tokenizer. Runs still optimizing are excluded,
since their argmax candidate can still change. Densities are per $1000$ tokens against these fixed
lexicons, matched case-insensitively:

\begin{itemize}\itemsep0.5pt
\item \textbf{Directives}: \emph{must}, \emph{must not}, \emph{do not}, \emph{don't}, \emph{never},
\emph{always}, \emph{exactly}, \emph{required}, \emph{ensure}, \emph{verify}, \emph{make sure},
\emph{you should}, \emph{be sure}, \emph{only}, \emph{only if}, \emph{important}, \emph{critical},
\emph{mandatory}, \emph{strictly}, \emph{precisely}.
\item \textbf{Prohibitions}: \emph{must not}, \emph{do not}, \emph{don't}, \emph{never},
\emph{avoid}, \emph{without}, \emph{no longer}, \emph{omit}, \emph{refrain}.
\item \textbf{Capitalized emphasis}: all-caps tokens of three or more letters.
\end{itemize}

All three are properties of the \emph{wording}. We do not count layout features (bullets, headings,
markup tags, backtick-quoted spans): the mechanism we hypothesize concerns how much a prompt spells
out, not how the text is arranged on the page, and a layout count would not test it.

The lexicons were fixed before the counts were inspected and are applied identically to the cheap,
full-cost and seed prompts, so they cannot favor either arm; they are nonetheless a crude proxy for
``explicitness,'' which is the main reason we read this table as descriptive rather than as a test of
the mechanism.

\paragraph{What the counts show.} We read the table as a set of \emph{paired} comparisons: each cheap
arm is compared only against the full-cost prompt of the tier it is deployed on, so the pairing holds
benchmark and deployment tier fixed and only the search configuration varies. Over the $52$ such pairs
($44$ distinct prompts spanning four benchmarks, five cheap arms and five full-cost tiers), the
cheap-search prompt is the longer of the two in $40$ pairs, with a median length ratio of
$1.29\times$. The differences are not only in length: normalised per $1000$ tokens, the cheap prompts
carry a median $1.26\times$ the directive density (higher in $35/52$ pairs), $1.17\times$ the
prohibition density ($34/52$), and $2.80\times$ the capitalised-emphasis density (higher in $36$ of the
$48$ pairs in which the full-cost prompt uses any capitalised emphasis at all; the four excluded pairs
are HoVer cells where that count is zero for the full-cost prompt and the ratio is undefined). All four
markers therefore move in the same direction: the prompts that transfer positively are the ones that
spell more out, as the structural-explicitness account of Section~\ref{sec:analysis} would predict.

\paragraph{Why we call this descriptive.} Three limits are worth stating plainly. First, this is a
correlation over the prompts the search happened to produce, not an ablation: we did not manipulate
explicitness and re-measure accuracy, so the counts cannot establish that explicitness is what carries
the transfer. Second, the ratios are medians over pairs that share prompts (a full-cost tier recurs
once per cheap arm compared against it), so the pairs are not independent and the ``higher in $k/n$''
tallies should be read as a description of the sign pattern, not as a test with a $p$-value. Third, a
lexicon count is a coarse instrument: it registers the vocabulary of explicit instruction, not whether
the instruction is well targeted. What the table does support is the weaker and still relevant claim
that the two arms produce systematically \emph{different} prompts rather than noisy variants of one
another, and that the difference has the shape the account predicts.

\input{tab_explicit}
\clearpage

\section{Per-Benchmark and Per-Family Results}
\label{app:detail}

This appendix holds the tables and the cell-by-cell discussion behind Section~\ref{sec:results}:
one table per model family over all four benchmarks, the shared neutral deploy target, and the
five numbered findings the main text summarizes.

\subsection{HotpotQA (easy task; lower-bound stress test)}
\label{sec:hotpot}
The HotpotQA section of Table~\ref{tab:fam_claude} shows that even where the optimizer
barely helps, the cheap
search matches the Haiku baseline when deployed on Haiku, at $\sim\!7\times$ lower
cost, and transfers further up to Sonnet.

\begin{table*}[t]\centering
\small\setlength{\tabcolsep}{4.5pt}
\begin{tabular}{llcccccc}
\toprule
\multicolumn{2}{c}{Method} & \multirow{2}{*}{opt cost \$} & \multirow{2}{*}{cost$\downarrow\times$} & \multirow{2}{*}{test @ nano} & \multirow{2}{*}{test @ mini$^{\ddagger}$} & \multirow{2}{*}{test @ Haiku} & \multirow{2}{*}{test @ Sonnet} \\
\cmidrule(lr){1-2}
Task model & Reflect model &  &  &  &  &  & \\
\midrule
\multicolumn{8}{l}{\emph{HotpotQA}} \\
\cmidrule(l){1-8}
Haiku & Haiku & $40.8$ & $1.0\times$ & -- & -- & $52.8\pm2.2$ & $58.3\pm2.0$ \\
mini$^{\S}$ & mini & $25.8$ & $1.6\times$ & -- & $52.2\pm3.7$ & $50.4\pm2.8$ & -- \\
nano & nano & $4.3$ & $9.5\times$ & $42.2\pm1.8$ & -- & $46.1\pm2.7$ & -- \\
nano & Haiku & $3.8$ & $10.8\times$ & $46.2\pm2.0$ & -- & $44.8\pm2.1$ & -- \\
\textbf{nano} & \textbf{Sonnet-5} & $\mathbf{6.0}$ & $\mathbf{6.8\times}$ & $47.4\pm2.2$ & $\mathbf{53.4\pm4.3}$ & $\mathbf{53.7\pm2.9}$ & $\mathbf{61.0\pm4.0}$ \\
\multicolumn{2}{l}{\emph{Baseline} (seed prompt)} & $0$ & -- & $31.7\pm3.2$ & $26.1\pm0.8$ & $29.2\pm2.5$ & $34.7\pm1.2$ \\
\midrule
\multicolumn{8}{l}{\emph{IFBench}} \\
\cmidrule(l){1-8}
Haiku & Haiku & $16.4$ & $1.0\times$ & -- & -- & $47.1\pm4.3$ & $72.4\pm0.6$ \\
mini$^{\S}$ & mini & $8.3$ & $2.0\times$ & -- & $53.2\pm2.5$ & $44.5\pm3.2$ & -- \\
\textbf{nano} & \textbf{Sonnet-5} & $\mathbf{2.4}$ & $\mathbf{6.9\times}$ & $40.2\pm1.3$ & $\mathbf{54.1\pm0.7}$ & $\mathbf{54.6\pm1.0}$ & $\mathbf{73.5\pm1.8}$ \\
\multicolumn{2}{l}{\emph{Baseline} (seed prompt)} & $0$ & -- & $37.4$ & $48.5$ & $21.3$ & $68.2$ \\
\midrule
\multicolumn{8}{l}{\emph{LiveBench-Math}} \\
\cmidrule(l){1-8}
Haiku & Haiku & $15.8$ & $1.0\times$ & -- & -- & $68.9\pm1.7$ & $\mathbf{66.0\pm1.1}^{\dagger}$ \\
mini$^{\S}$ & mini & $5.9$ & $2.7\times$ & -- & $\mathbf{60.1\pm0.9}$ & $67.9\pm0.6$ & -- \\
\textbf{nano} & \textbf{Sonnet-5} & $\mathbf{2.6}$ & $\mathbf{6.0\times}$ & $45.7\pm1.3$ & $\mathbf{60.1\pm1.7}$ & $\mathbf{69.5\pm3.3}$ & $61.1\pm4.0$ \\
\multicolumn{2}{l}{\emph{Baseline} (seed prompt)} & $0$ & -- & $41.5$ & $58.8$ & $63.0$ & $63.5$ \\
\midrule
\multicolumn{8}{l}{\emph{HoVer}} \\
\cmidrule(l){1-8}
Haiku & Haiku & $81.3$ & $1.0\times$ & -- & -- & $\mathbf{62.1\pm1.0}$ & $65.8\pm3.6^{\dagger}$ \\
mini$^{\S}$ & mini & $25.1$ & $3.2\times$ & -- & $51.4\pm0.8$ & $54.8\pm3.0$ & $61.8\pm1.2$ \\
\textbf{nano} & \textbf{Sonnet-5} & $\mathbf{10.2}$ & $\mathbf{8.0\times}$ & $51.0\pm0.6$ & $\mathbf{55.2\pm1.7}$ & $58.3\pm0.9$ & $\mathbf{67.3\pm0.9}$ \\
\multicolumn{2}{l}{\emph{Baseline} (seed prompt)} & $0$ & -- & $34.3$ & $42.7$ & $48.0$ & $55.0$ \\
\bottomrule
\end{tabular}
\caption{\textbf{Mixed Claude family} (Section~\ref{sec:setup}): nano answering with a Sonnet-5 reflector,
deployed across a four-tier ladder, on all four benchmarks. Conventions as in
Table~\ref{tab:fam_gpt}. $^{\ddagger}$\emph{test @ mini} is a neutral cross-family mid tier that is
not the reflector (Table~\ref{tab:mini_neutral}); $^{\S}$Full-mini is \texttt{gpt-4.1-mini} optimized
on its own tier, from Table~\ref{tab:fam_gpt}; $^{\dagger}$cross-tier deploy of the full
Haiku-optimized prompt on Sonnet. \emph{test @ Sonnet} cells are re-scored at a uniform high token cap.
The two weak-reflector rows (the evidence for Finding~1) were run on HotpotQA only.}
\label{tab:fam_claude}
\end{table*}

\noindent\emph{Finding 1 (reflector strength is the lever).} Nano+\emph{Haiku}-refl
deployed on Haiku ($44.8$) is no better than plain Full-Nano ($46.1$); only the
\emph{strong} (Sonnet) reflector closes the gap. Cheapness alone is insufficient: the
operator must be strong.

\paragraph{The operator premium is deliberate.} The reflector is rare in calls and tokens, at most a
few percent of each, yet its per-token price is an order of magnitude above the cheap answerer's, so it
still carries roughly a third of the optimization bill. We pay that knowingly: Finding~1 shows a cheap
reflector does not produce transferable prompts, so the operator is where strength must be bought. The
saving comes from the evaluation role, which holds the other ${>}96\%$ of tokens, and it is large enough
that the net reduction stays in the $5.6$--$14\times$ range. The call and token shares are price-free
measurements; the dollar split is what a future price list would move, so we report both, and
Appendix~\ref{app:breakeven} gives the point at which the trade would stop paying.

\subsection{IFBench (harder task; optimizer has headroom)}

\noindent\emph{Finding 2 (cheap search wins on the harder task).} On IFBench the cheap prompt beats
Full-Haiku on Haiku by $7.5$ points at $\sim\!7\times$ lower cost, with lower variance, and keeps the
lead when both prompts move up to Sonnet.

\subsection{Model-pair generalization and tier portability (Azure GPT-5.x)}
The IFBench section of Table~\ref{tab:fam_gpt} is the headline: one cheap search deployed
on each tier vs.\ that tier's own full-cost optimization. That table gathers all four
benchmarks for this family, one section each, over the same three tiers.

\begin{table*}[t]\centering
\small\setlength{\tabcolsep}{4.5pt}
\begin{tabular}{llccccc}
\toprule
\multicolumn{2}{c}{Method} & \multirow{2}{*}{opt cost \$} & \multirow{2}{*}{cost$\downarrow\times$} & \multirow{2}{*}{test @ nano} & \multirow{2}{*}{test @ mini} & \multirow{2}{*}{test @ luna} \\
\cmidrule(lr){1-2}
Task model & Reflect model &  &  &  &  & \\
\midrule
\multicolumn{7}{l}{\emph{HotpotQA}} \\
\cmidrule(l){1-7}
gpt-5.6-luna & gpt-5.6-luna & $102.0$ & $1.0\times$ & -- & -- & $50.4\pm2.6$ \\
gpt-4.1-mini & gpt-4.1-mini & $25.8$ & $3.9\times$ & -- & $52.2\pm3.7$ & $52.2\pm4.0^{\dagger}$ \\
\textbf{nano} & \textbf{gpt-5.5} & $\mathbf{7.3}$ & $\mathbf{14.1\times}$ & $47.7\pm2.2$ & $\mathbf{52.3\pm4.2}$ & $\mathbf{59.4\pm1.2}$ \\
\multicolumn{2}{l}{\emph{Baseline} (seed prompt)} & $0$ & -- & $31.7\pm3.2$ & $26.1\pm0.8$ & $39.6\pm1.3$ \\
\midrule
\multicolumn{7}{l}{\emph{IFBench}} \\
\cmidrule(l){1-7}
gpt-5.6-luna & gpt-5.6-luna & $32.6$ & $1.0\times$ & -- & -- & $66.6\pm2.2$ \\
gpt-4.1-mini & gpt-4.1-mini & $8.3$ & $3.9\times$ & -- & $\mathbf{53.2\pm2.5}$ & $66.7\pm1.9^{\dagger}$ \\
\textbf{nano} & \textbf{gpt-5.5} & $\mathbf{3.0}$ & $\mathbf{11.0\times}$ & $40.6\pm2.9$ & $51.0\pm2.8$ & $\mathbf{67.9\pm1.9}$ \\
\multicolumn{2}{l}{\emph{Baseline} (seed prompt)} & $0$ & -- & $37.4$ & $48.5$ & $64.3$ \\
\midrule
\multicolumn{7}{l}{\emph{LiveBench-Math}} \\
\cmidrule(l){1-7}
gpt-5.6-luna & gpt-5.6-luna & $16.3$ & $1.0\times$ & -- & -- & $\mathbf{76.4\pm2.1}$ \\
gpt-4.1-mini & gpt-4.1-mini & $5.9$ & $2.8\times$ & -- & $\mathbf{60.1\pm0.9}$ & $75.7\pm4.1^{\dagger}$ \\
\textbf{nano} & \textbf{gpt-5.5} & $\mathbf{2.9}$ & $\mathbf{5.6\times}$ & $44.0\pm6.3$ & $59.2\pm1.1$ & $75.2\pm1.1$ \\
\multicolumn{2}{l}{\emph{Baseline} (seed prompt)} & $0$ & -- & $41.5$ & $58.8$ & $72.1$ \\
\midrule
\multicolumn{7}{l}{\emph{HoVer}} \\
\cmidrule(l){1-7}
gpt-5.6-luna & gpt-5.6-luna & $94.2$ & $1.0\times$ & -- & -- & $\mathbf{64.8\pm0.5}$ \\
gpt-4.1-mini & gpt-4.1-mini & $25.1$ & $3.7\times$ & -- & $\mathbf{51.4\pm0.8}$ & $60.1\pm2.6^{\dagger}$ \\
\textbf{nano} & \textbf{gpt-5.5} & $\mathbf{8.5}$ & $\mathbf{11.1\times}$ & $48.8\pm3.2$ & $49.8\pm1.0$ & $63.6\pm0.5$ \\
\multicolumn{2}{l}{\emph{Baseline} (seed prompt)} & $0$ & -- & $33.0$ & $42.7$ & $51.7$ \\
\bottomrule
\end{tabular}
\caption{\textbf{GPT family} (Azure): one cheap search deployed on each tier against that tier's own
full-cost optimization, on all four benchmarks. \% task metric, mean$\pm$sample std over $n{=}3$ seeds
(the convention in every table here); opt cost is the mean \$ from the token logs at the prices of
Table~\ref{tab:prices}; cost$\downarrow\times$ is against the costliest run within that benchmark.
Full-$X$ optimizes on tier $X$; the cheap combo searches once with nano answering and is deployed on
each tier. \emph{test @ nano} is its own search score, so the rise from there to \emph{test @ luna} is
the deploy model's own capability rather than a transfer gain (the $G$ vs.\ $R$ distinction of
Section~\ref{sec:problem}). Best mean per deploy column in bold; ``--'' = not run on that tier.
Baseline rows carry a $\pm$ on HotpotQA only, the one benchmark whose test split is re-drawn per seed.
$^{\dagger}$cross-tier deploy of the full mini-optimized prompt on luna.}
\label{tab:fam_gpt}
\end{table*}

The same GPT-family cross-tier comparison on HotpotQA (HotpotQA section of
Table~\ref{tab:fam_gpt})
confirms the effect on a second task: the cheap nano+gpt-5.5 search matches Full-mini on mini and
beats Full-luna on luna by $9$ points at $\sim\!14\times$ lower cost.

\noindent\emph{Finding 3 (tier portability).} On IFBench one $\sim\!\$3$ prompt matches each tier's
own full optimization, including the \$33 Luna run at $\sim\!11\times$ lower cost, with regret
$R_{\text{nano}\to\text{luna}}=-1.3$. The same prompt rises $\sim\!27$ points from nano to Luna, which
is the deploy model's own capability rather than a transfer gain, since the untrained seed prompt makes
the same jump. The load-bearing claim is the near-zero regret.

Upward transfer is not specific to a cheap search: a prompt optimized end-to-end on gpt-4.1-mini also
gains $+14$ points deployed on Luna and matches Full-Luna there. The nano combos reach the same place
for a fraction of the search cost.

\subsection{LiveBench-Math (math reasoning, a low-headroom regime)}
\label{sec:lbmath}
LiveBench-Math adds math reasoning, exactly per the GEPA-artifact setup. The recipe again transfers
upward and stays comparable to full same-tier optimization at a fraction of the cost. Here every
method, including the \$0 seed prompt, lands within a few points on each tier, which the
zero-optimization control makes explicit (Finding~4).

The Mixed Claude family shows the same pattern with a tier-dependent split. Haiku keeps some
headroom, so the cheap prompt matches Full-Haiku there and both clear the seed by several points.
Sonnet is near saturated from the seed alone, and there the cheap prompt transfers upward less
cleanly.

\subsection{HoVer (3-hop claim-verification retrieval)}
HoVer is retrieval-only (binary 3-hop recall, no answer LLM and no judge; the faithful gepa-artifact
3-hop subset, 150/300/300), so it stresses multi-hop query formulation rather than answering. The
pattern holds. The cheap nano+Sonnet prompt transfers upward across all three tiers, clears the \$0
seed floor at each by $10$ points or more, lands within $\sim\!4$ points of Full-Haiku on Haiku at
$\sim\!8\times$ lower cost, and on Sonnet the \$10 prompt exceeds the \$81 Full-Haiku prompt.

\subsection{HoVer (GPT family)}
The GPT-family HoVer runs repeat the recipe on a second family and give the sharpest cost story. The
cheap prompt deployed upward on luna comes within $1.2$ points of Full-luna at $11\times$ lower cost,
and beats the costlier Full-mini prompt on the same tier.
Full-luna optimizes HoVer well (best-val grew $+0.15$ over the seed), so the near-match is a real
result rather than a low ceiling.

\begin{table*}[t]\centering
\small\setlength{\tabcolsep}{4.5pt}
\begin{tabular}{llccccc}
\toprule
\multicolumn{2}{c}{Method} & \multirow{2}{*}{opt cost \$} & \multirow{2}{*}{cost$\downarrow\times$} & \multirow{2}{*}{test @ flash-lite} & \multirow{2}{*}{test @ 3.5-flash} & \multirow{2}{*}{test @ 2.5-pro} \\
\cmidrule(lr){1-2}
Task model & Reflect model &  &  &  &  & \\
\midrule
\multicolumn{7}{l}{\emph{HotpotQA}} \\
\cmidrule(l){1-7}
2.5-pro & 2.5-pro & $564.5$ & $1.0\times$ & -- & -- & $56.6\pm4.2$ \\
3.5-flash & 3.5-flash & $467.4$ & $1.2\times$ & -- & $58.8\pm0.5$ & -- \\
\textbf{flash-lite} & \textbf{3.1-pro} & $\mathbf{10.5}$ & $\mathbf{53.9\times}$ & $50.1\pm2.8$ & $\mathbf{59.1\pm3.3}$ & $\mathbf{58.0\pm2.7}$ \\
\multicolumn{2}{l}{\emph{Baseline} (seed prompt)} & $0$ & -- & $35.2\pm1.9$ & $49.1\pm0.2$ & $23.4\pm1.5$ \\
\midrule
\multicolumn{7}{l}{\emph{IFBench}} \\
\cmidrule(l){1-7}
2.5-pro & 2.5-pro & $182.9$ & $1.5\times$ & -- & -- & $67.5\pm1.1$ \\
3.5-flash & 3.5-flash & $274.8$ & $1.0\times$ & -- & $\mathbf{71.2\pm3.4}$ & -- \\
\textbf{flash-lite} & \textbf{3.1-pro} & $\mathbf{7.4}$ & $\mathbf{37.3\times}$ & $53.0\pm1.3$ & $69.7\pm1.7$ & $\mathbf{67.7\pm1.9}$ \\
\multicolumn{2}{l}{\emph{Baseline} (seed prompt)} & $0$ & -- & $48.1$ & $73.1$ & $63.6$ \\
\midrule
\multicolumn{7}{l}{\emph{LiveBench-Math}} \\
\cmidrule(l){1-7}
2.5-pro & 2.5-pro & $269.9$ & $1.0\times$ & -- & -- & $76.4\pm0.4$ \\
3.5-flash & 3.5-flash & $97.0$ & $2.8\times$ & -- & $78.9\pm2.0$ & -- \\
\textbf{flash-lite} & \textbf{3.1-pro} & $\mathbf{10.6}$ & $\mathbf{25.5\times}$ & $63.2\pm2.0$ & $\mathbf{81.3\pm1.3}$ & $\mathbf{76.9\pm0.2}$ \\
\multicolumn{2}{l}{\emph{Baseline} (seed prompt)} & $0$ & -- & $60.9$ & $73.9$ & $76.4$ \\
\midrule
\multicolumn{7}{l}{\emph{HoVer}} \\
\cmidrule(l){1-7}
2.5-pro & 2.5-pro & $491.0$ & $1.0\times$ & -- & -- & $59.0\pm2.3$ \\
3.5-flash & 3.5-flash & $331.8$ & $1.5\times$ & -- & $\mathbf{70.3\pm1.7}$ & -- \\
\textbf{flash-lite} & \textbf{3.1-pro} & $\mathbf{11.4}$ & $\mathbf{42.9\times}$ & $54.7\pm2.3$ & $68.9\pm6.6$ & $\mathbf{64.2\pm2.5}$ \\
\multicolumn{2}{l}{\emph{Baseline} (seed prompt)} & $0$ & -- & $43.3$ & $40.7$ & $49.7$ \\
\bottomrule
\end{tabular}
\caption{\textbf{Gemini family} (GCP), the third vendor: \texttt{gemini-2.5-flash-lite} answering
with a \texttt{gemini-3.1-pro} reflector, deployed up on \texttt{gemini-3.5-flash} and
\texttt{gemini-2.5-pro}, on all four benchmarks. Conventions as in Table~\ref{tab:fam_gpt}. The cost
asymmetry is largest in this family, $25$--$54\times$, because both Gemini reasoning tiers emit long
chains of thought during evaluation. All \texttt{2.5-pro} cells are re-scored at a uniform high token
cap with zero residual generation failures. ``--'' = not run on that tier.}
\label{tab:fam_gemini}
\end{table*}

\begin{table*}[t]\centering
\small\setlength{\tabcolsep}{4.5pt}
\begin{tabular}{llcccccc}
\toprule
\multicolumn{2}{c}{Method} & \multirow{2}{*}{opt cost \$} & \multirow{2}{*}{cost$\downarrow\times$} & \multirow{2}{*}{test @ Qwen3-8B} & \multirow{2}{*}{test @ mini} & \multirow{2}{*}{test @ Haiku} & \multirow{2}{*}{test @ luna} \\
\cmidrule(lr){1-2}
Task model & Reflect model &  &  &  &  &  & \\
\midrule
\multicolumn{8}{l}{\emph{HotpotQA}} \\
\cmidrule(l){1-8}
Haiku & Haiku & $40.8$ & $2.5\times$ & -- & -- & $52.8\pm2.2$ & -- \\
mini & mini & $25.8$ & $3.9\times$ & -- & $52.2\pm3.7$ & $50.4\pm2.8^{\dagger}$ & $52.2\pm4.0^{\dagger}$ \\
luna & luna & $102.0$ & $1.0\times$ & -- & -- & -- & $50.4\pm2.6$ \\
\textbf{Qwen3-8B} & \textbf{Sonnet-5} & $\mathbf{1.6}$ & $\mathbf{63.4\times}$ & $49.3\pm3.8$ & $55.1\pm1.4$ & $53.7\pm3.5$ & $\mathbf{55.3\pm2.0}$ \\
\textbf{Qwen3-8B} & \textbf{gpt-5.5} & $\mathbf{0.9}$ & $\mathbf{114.2\times}$ & $48.0\pm2.6$ & $\mathbf{55.6\pm1.6}$ & $\mathbf{55.1\pm1.7}$ & $54.0\pm5.2$ \\
\midrule
\multicolumn{8}{l}{\emph{IFBench}} \\
\cmidrule(l){1-8}
Haiku & Haiku & $16.4$ & $2.0\times$ & -- & -- & $47.1\pm4.3$ & -- \\
mini & mini & $8.3$ & $3.9\times$ & -- & $53.2\pm2.5$ & $44.5\pm3.2^{\dagger}$ & $66.7\pm1.9^{\dagger}$ \\
luna & luna & $32.6$ & $1.0\times$ & -- & -- & -- & $66.6\pm2.2$ \\
\textbf{Qwen3-8B} & \textbf{Sonnet-5} & $\mathbf{0.7}$ & $\mathbf{47.0\times}$ & $45.9\pm2.9$ & $\mathbf{53.7\pm0.4}$ & $52.4\pm3.0$ & $\mathbf{67.5\pm2.5}$ \\
\textbf{Qwen3-8B} & \textbf{gpt-5.5} & $\mathbf{0.9}$ & $\mathbf{36.4\times}$ & $46.1\pm4.3$ & $50.9\pm1.6$ & $\mathbf{53.8\pm1.4}$ & $67.4\pm3.5$ \\
\midrule
\multicolumn{8}{l}{\emph{LiveBench-Math}} \\
\cmidrule(l){1-8}
Haiku & Haiku & $15.8$ & $1.0\times$ & -- & -- & $68.9\pm1.7$ & -- \\
mini & mini & $5.9$ & $2.8\times$ & -- & $60.1\pm0.9$ & $67.9\pm0.6^{\dagger}$ & $75.7\pm4.1^{\dagger}$ \\
luna & luna & $16.3$ & $1.0\times$ & -- & -- & -- & $76.4\pm2.1$ \\
\textbf{Qwen3-8B} & \textbf{Sonnet-5} & $\mathbf{1.1}$ & $\mathbf{15.2\times}$ & $67.5\pm0.6$ & $\mathbf{61.1\pm1.2}$ & $\mathbf{69.9\pm3.3}$ & $\mathbf{77.2\pm2.2}$ \\
\textbf{Qwen3-8B} & \textbf{gpt-5.5} & $\mathbf{1.0}$ & $\mathbf{15.8\times}$ & $70.2\pm1.5$ & $60.1\pm0.8$ & $68.1\pm2.2$ & $76.1\pm1.9$ \\
\midrule
\multicolumn{8}{l}{\emph{HoVer}} \\
\cmidrule(l){1-8}
Haiku & Haiku & $81.3$ & $1.2\times$ & -- & -- & $62.1\pm1.0$ & -- \\
mini & mini & $25.1$ & $3.7\times$ & -- & $51.4\pm0.8$ & $54.8\pm3.0^{\dagger}$ & $60.1\pm2.6^{\dagger}$ \\
luna & luna & $94.2$ & $1.0\times$ & -- & -- & -- & $64.8\pm0.5$ \\
\textbf{Qwen3-8B} & \textbf{Sonnet-5} & $\mathbf{2.0}$ & $\mathbf{47.8\times}$ & $53.8\pm2.0$ & $\mathbf{58.0\pm1.2}$ & $\mathbf{64.0\pm2.0}$ & $\mathbf{65.9\pm5.8}$ \\
\textbf{Qwen3-8B} & \textbf{gpt-5.5} & $\mathbf{1.7}$ & $\mathbf{54.6\times}$ & $54.6\pm2.3$ & $56.8\pm0.5$ & $61.6\pm0.8$ & $65.8\pm1.0$ \\
\bottomrule
\end{tabular}
\caption{\textbf{Zero-API-cost answerer limit} (Qwen3-8B): the answering role moves to a self-hosted
open-weights model with no per-token charge, so the whole optimization cost is the paid reflector's, and
the evolved prompt is deployed up on three paid tiers. The three Full-$X$ rows are the same own-tier
optimizations as Tables~\ref{tab:fam_gpt} and~\ref{tab:fam_claude}, repeated as the reference for these
rows. Conventions as in Table~\ref{tab:fam_gpt}. \emph{test @ Qwen} is the local answerer's own search
score. $^{\dagger}$cross-tier deploy of the full mini-optimized prompt. Analysis in
Appendix~\ref{sec:qwen}.}
\label{tab:fam_qwen}
\end{table*}

\noindent\emph{Finding 4 (a zero-optimization baseline shows where headroom exists).} We deploy the
untrained seed prompt (candidate~\#0, cost \$0) on every tier and report it as a row in every table.
Where optimization has headroom the optimized prompts sit far above that floor, by $12$--$26$ points on
HotpotQA and by $+33$ on IFBench@Haiku. On tiers already near their prompt-insensitive ceiling the gap
closes: LiveBench-Math and IFBench@luna reach within a few points of every optimized method from the
seed alone. This scopes the claim. Cost-aware transfer captures large gains where prompt optimization
matters, and where it does not the cheap recipe still matches full-cost optimization. On these four
tasks we did not observe it fall below full same-tier optimization. The search curves agree: validation
climbs far less on LiveBench-Math than on HotpotQA and IFBench.

\subsection{A third vendor: the Gemini family}
The two families above share the same \texttt{gpt-4.1-nano} answerer, so the result could be read as a
property of that one model. Table~\ref{tab:fam_gemini} repeats the whole protocol inside a third
vendor's ladder with no OpenAI or Anthropic model involved, on all four benchmarks at $n{=}3$ seeds.

The recipe holds and the cost asymmetry is much larger. On \texttt{2.5-pro} the cheap prompt matches or
beats that tier's own full-cost optimization on all four benchmarks, for under \$$12$ against
\$$183$--\$$564$, a $25$--$54\times$ reduction. On \texttt{3.5-flash} it wins on two benchmarks and
trails by at most $1.5$ points on the other two. The larger multiplier is mechanical: both Gemini reasoning tiers
emit long chains of thought on every fitness evaluation, so moving the high-volume answering role off
them saves proportionally more. This is the regime the cost model of Section~\ref{sec:problem} predicts
the method exploits best, since the saving grows with the per-token gap between the two tiers.

The Gemini runs sharpen Finding~4 in both directions. On HotpotQA the seed prompt is weakest on the
\emph{strongest} model, which is the least willing to emit a bare short-form answer, so optimization is
worth $+34.6$ points there. On IFBench@$3.5$-flash the \$$0$ seed prompt scores above every optimized
method on that tier. That is the clearest negative instance in the paper, and we report it rather than
dropping the cell.

\subsection{Pushing the answerer to \$0: a self-hosted open-weights searcher}
\label{sec:qwen}
If the answering role only needs to rank candidates, the limit of component (A) is an answerer that
costs nothing per token. Table~\ref{tab:fam_qwen} takes it there. The answerer is a self-hosted,
open-weights Qwen3-8B, so the entire optimization bill is the rare reflector's, under \$$2$ per run,
and the evolved prompt is deployed up on three paid tiers. The Full-$X$ rows repeated in that table are
the same own-tier optimizations as Tables~\ref{tab:fam_gpt} and~\ref{tab:fam_claude}, so the
zero-API-cost rows read directly against them. Every dollar figure here is API spend; the serving
compute is out of scope.

A prompt searched for under \$$2$ by a local 8B model matches or beats every paid tier's own full-cost
optimization. HotpotQA gives the largest reduction, $63$--$114\times$, with a gain on all three tiers
under either reflector. IFBench and LiveBench-Math match their own-tier references at $15$--$47\times$
less, though LiveBench-Math is the low-headroom task of Appendix~\ref{sec:lbmath}, so read those
margins as not collapsing rather than as a gain. The two reflectors land on top of each other at almost
every tier, at the opposite end of the answerer price range: the loop needs a strong operator and a
discriminative evaluator rather than an accurate one.

HoVer is the one task where the two reflectors separate. At $48$--$55\times$ less, five of the six
deployed cells are at or above their own-tier reference and the sixth falls $0.6$ points short, inside
the seed spread on both sides. We therefore state HoVer as matching or beating own-tier optimization on
five of six cells and tying on the sixth. The reflector gap here, with Sonnet-5 ahead on mini and
Haiku, is the one place in this table where operator strength shows in the deployed number.

Two details of the HoVer rows matter for anyone reading a single column. The local answerer's own
search score ranks the two reflectors the wrong way round: gpt-5.5 leads during search and loses at two
deploy tiers, so only the deployed column tracks transfer quality. And the two luna means agree to
$0.1$ points while their spreads differ sixfold, the wide one from a single low seed; an independent
reflector arm reproduces the mean, so we read that seed as ordinary variance.

\subsection{A shared neutral deploy target: cross-reflector-family transfer}
The Mixed Claude family's costliest deploy tier doubles as its reflector, so ``deploy on Sonnet'' is not a fully
unseen-model test. The two families share the identical \texttt{gpt-4.1-nano} answerer and differ only
in the reflector, so we deploy the Mixed Claude prompt on \texttt{gpt-4.1-mini}, a mid model the Sonnet
reflector never saw, beside the GPT-family cheap prompt and full same-tier optimization on that same
target (Table~\ref{tab:mini_neutral}).

\begin{table*}[h!]\centering
\small\setlength{\tabcolsep}{5pt}
\begin{tabular}{@{}llcccc@{}}
\toprule
Prompt source (deployed on gpt-4.1-mini) & reflector & HotpotQA & IFBench & LiveB.-Math & HoVer \\
\midrule
Full-mini (own-tier optimization) & --        & $52.2\pm3.7$ & $53.2\pm2.5$ & $60.1\pm0.9$ & $51.4\pm0.8$ \\
Cheap nano-search (GPT family)   & gpt-5.5   & $52.3\pm4.2$ & $51.0\pm2.8$ & $59.2\pm1.1$ & $49.8\pm1.0$ \\
\textbf{Cheap nano-search (Mixed Claude)} & \textbf{Sonnet-5} & $\mathbf{53.4\pm4.3}$ & $\mathbf{54.1\pm0.7}$ & $\mathbf{60.1\pm1.7}$ & $\mathbf{55.2\pm1.7}$ \\
\bottomrule
\end{tabular}
\caption{Cross-reflector-family transfer on a shared neutral deploy target (\texttt{gpt-4.1-mini}).
All three rows are scored on the same model; the two cheap rows share the nano answerer and differ only
in the reflector. Mean$\pm$sample std, $n{=}3$. The Sonnet-reflected prompt matches or beats both the
gpt-5.5-reflected prompt and full same-tier mini optimization.}
\label{tab:mini_neutral}
\end{table*}

\noindent\emph{Finding 5 (transfer is reflector-family agnostic).} A prompt evolved with an Anthropic
reflector and answered by nano deploys onto an OpenAI mid model indistinguishably from the same recipe
with a gpt-5.5 reflector and from full mini-only optimization, on every task. The gain rests on the
weak-answerer / strong-reflector structure rather than on any affinity between reflector and deploy
model, and it holds on a genuinely unseen mid tier.

\newpage

\section{The Pooled Transfer Residual}
\label{app:residual}

\begin{figure}[]\centering
\includegraphics[width=\columnwidth]{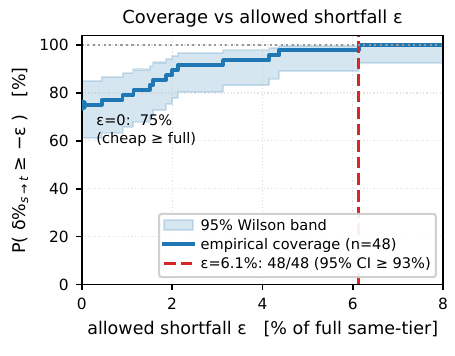}
\caption{Coverage of the same $48$ residuals: the fraction of setups landing within $\varepsilon\%$ of full same-tier optimization, with a $95\%$ Wilson band.}
\label{fig:coverage}
\end{figure}

This appendix gives the setups, statistics and reading caveats behind the pooled residual of
Section~\ref{sec:analysis} and Figures~\ref{fig:forest}--\ref{fig:coverage}.

\paragraph{The twelve setups.} Two GPT ladders (cheap$\to$mini, cheap$\to$luna), two Mixed Claude
ladders (cheap$\to$mini, cheap$\to$Haiku), two Gemini ladders (flash-lite$\to$3.5-flash and
$\to$2.5-pro) and six self-hosted Qwen ladders (two reflectors $\times$ mini, Haiku and luna). Same
protocol, three seeds and the same token caps throughout. Each setup contributes one residual per
dataset, the cheap cross-tier score minus the matching full same-tier score, divided by
$J_t(\pi_t^\ast)$ so the values are scale-free. That gives $48$ values. The residual is the negated
target regret of Section~\ref{sec:problem}; we negate so that higher is better.

›

\paragraph{Where the distribution sits.}
The pooled mean is $\delta^{\%}={+}2.8\%$, $95\%$ CI $[{+}1.3\%,{+}4.4\%]$, and the cheap prompt matches or beats the expensive one in $36$ of $48$ setups. Per family the means run ${+}1.0\%$ (GPT), ${+}1.5\%$ (Gemini), ${+}3.0\%$ (Mixed Claude) and ${+}3.8\%$ (Mixed Qwen). All four intervals overlap, so the families are indistinguishable at this sample size, but the ordering is the one the mechanism suggests: Mixed Qwen, the widest search/deploy capability gap we test, is the most favourable and the only family whose interval excludes zero. Mixed Qwen contributes half the sample, so the figure shows per-family rows rather than one pooled density.

\paragraph{The coverage reading.} Figure~\ref{fig:coverage} answers a practitioner's question: if I
allow the cheap prompt to fall $\varepsilon\%$ short of full same-tier optimization, how often does it
clear that bar? At zero tolerance, $36/48$ setups ($75\%$, $95\%$ Wilson $[61\%,85\%]$). Allowing a
$2\%$ shortfall, $43/48$ ($90\%$). The curve reaches $48/48$ at $\varepsilon=6.12\%$, which is simply
the largest shortfall we observed, so read it as a descriptive bound on this sample rather than a
prediction: the worst of these $48$ transfers gave up $6.12\%$, and doubling the sample left that worst
case unchanged. The upside tail reaches ${+}17.9\%$, so the residuals are right-skewed. We plot the
empirical coverage rather than a smoothed version, which would place mass below the smallest residual
we saw. One caveat: a variance-weighted mean, which down-weights the noisier large-margin points, sits
slightly below zero, so the centre is within about a point of zero either way. The shape is what is
robust, a distribution concentrated near zero with a bounded downside and a heavier upside.

\section{Validation Curves During Search}
\label{app:curves}

\begin{figure*}[t]\centering
\includegraphics[width=\linewidth]{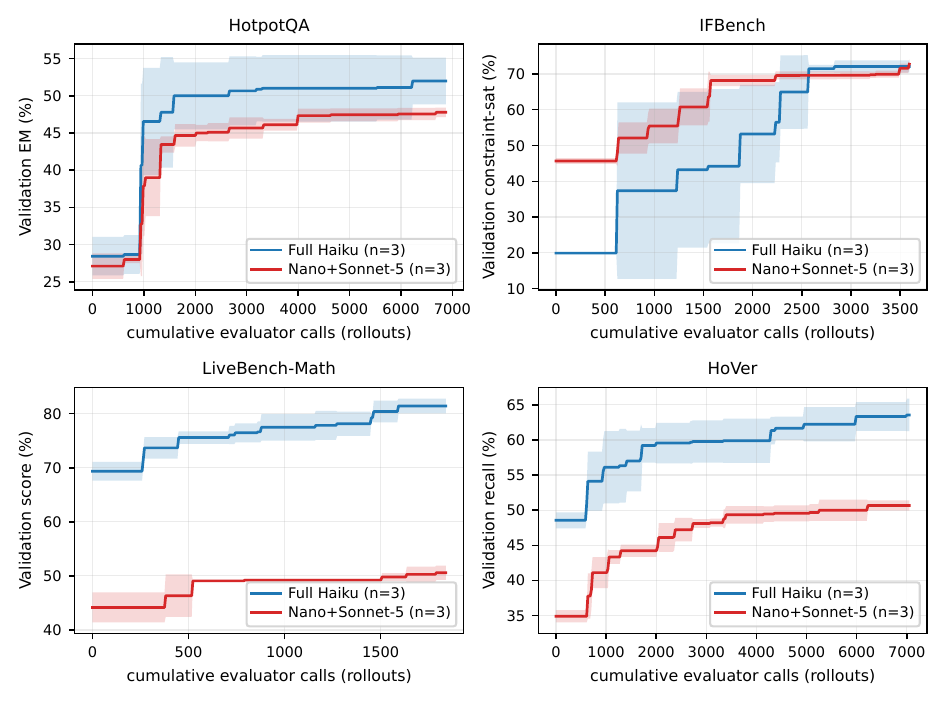}
\caption{Cross-vendor family (\texttt{nano} answerer, Sonnet-5 reflector) against full-cost Haiku
search, four benchmarks. Mean over $3$ seeds, $\pm1$ std.}
\label{fig:curves_crossvendor}
\end{figure*}

\begin{figure*}[t]\centering
\includegraphics[width=\linewidth]{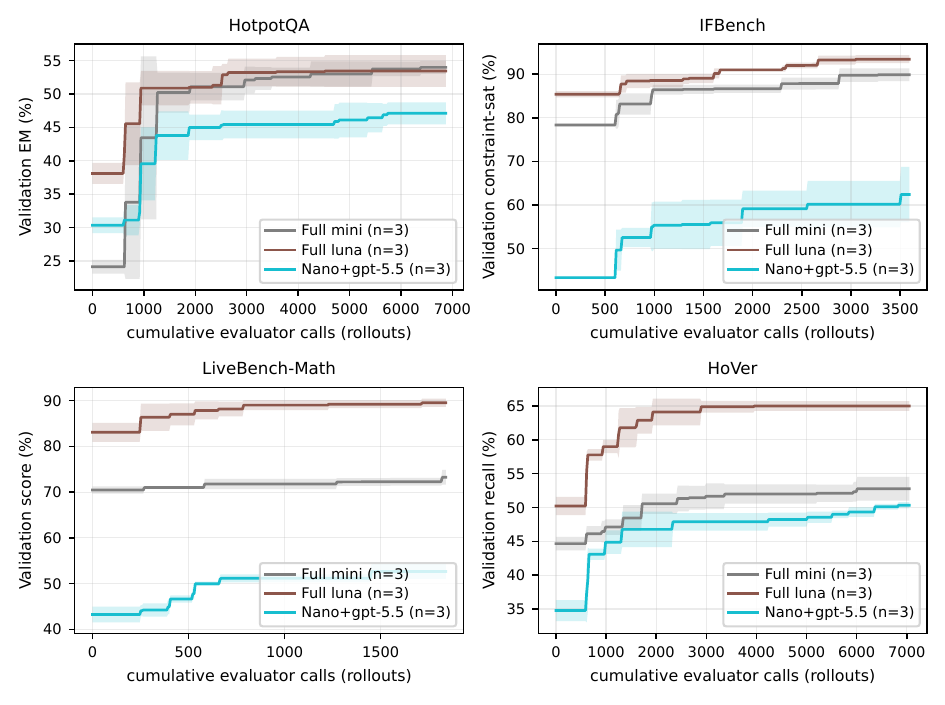}
\caption{GPT family: cheap \texttt{nano}$+$\texttt{gpt-5.5} search against full-cost \texttt{mini} and
\texttt{luna} search, four benchmarks. Mean over $3$ seeds, $\pm1$ std.}
\label{fig:curves_gpt}
\end{figure*}

\begin{figure*}[t]\centering
\includegraphics[width=\linewidth]{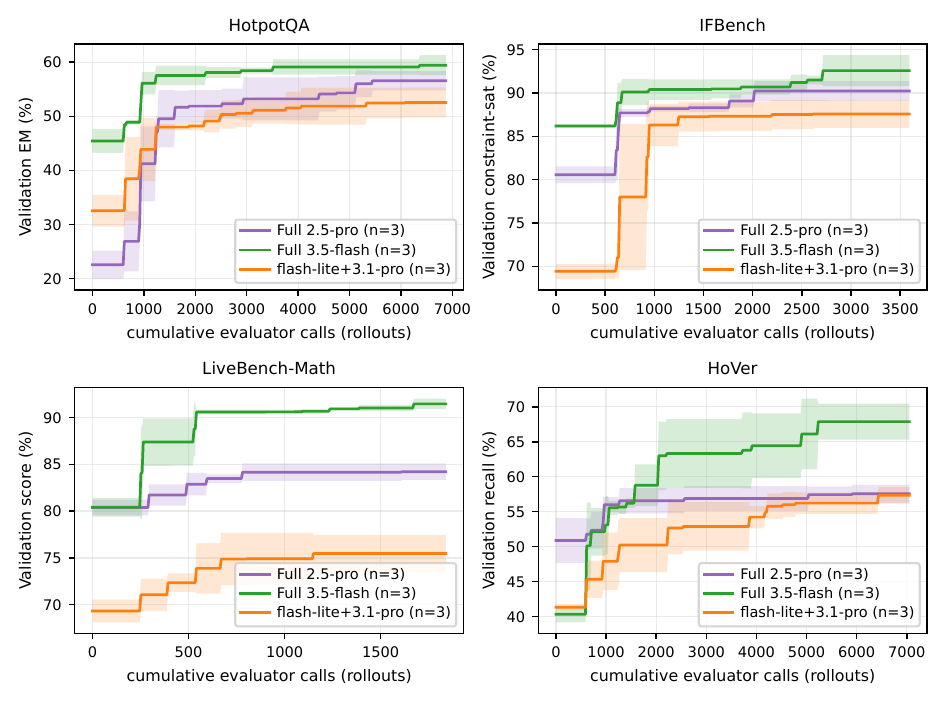}
\caption{Gemini family: cheap \texttt{flash-lite}$+$\texttt{3.1-pro} search against full-cost
\texttt{3.5-flash} and \texttt{2.5-pro} search, four benchmarks. Mean over $3$ seeds, $\pm1$ std.}
\label{fig:curves_gemini}
\end{figure*}

\begin{figure*}[t]\centering
\includegraphics[width=\linewidth]{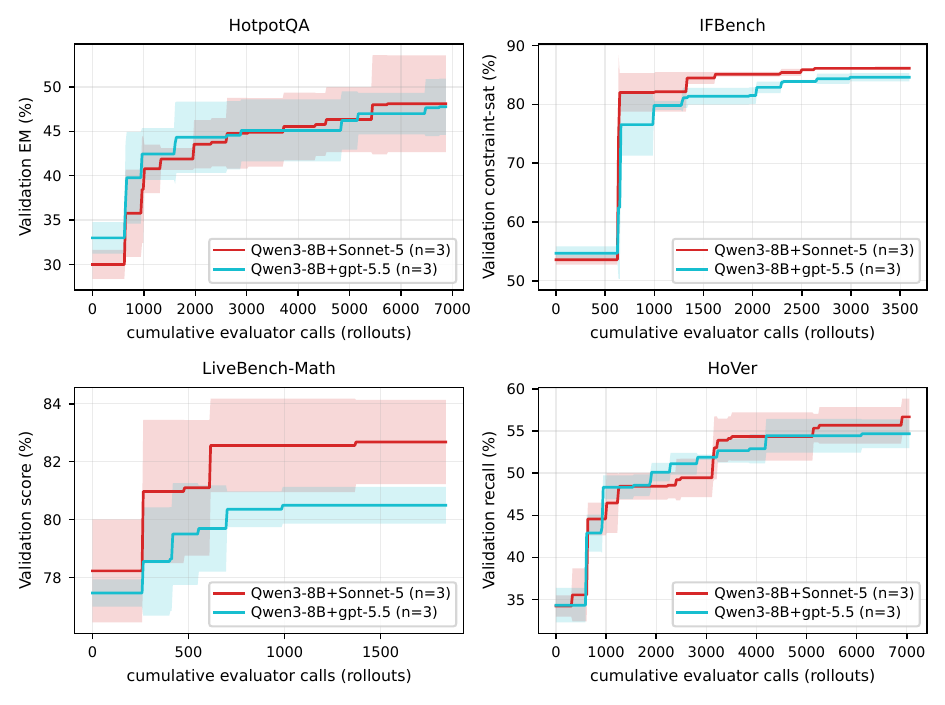}
\caption{Mixed Qwen family: a self-hosted \texttt{Qwen3-8B} answerer with two paid reflectors, four
benchmarks. No full-cost arm exists here, as the answerer carries no API cost at any point. Mean over $3$ seeds, $\pm1$ std.}
\label{fig:curves_qwen}
\end{figure*}

\begin{figure*}[t]\centering
\includegraphics[width=\linewidth]{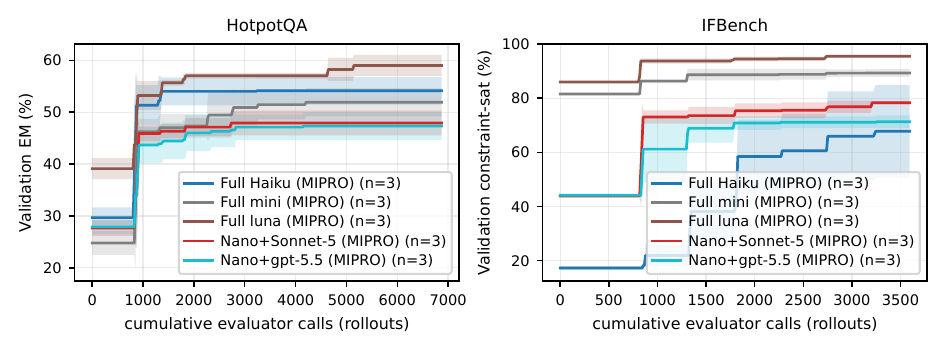}
\caption{MIPROv2 ablation: three full-cost and two cheap arms, on the two benchmarks it was run on.
Mean over $3$ seeds, $\pm1$ std.}
\label{fig:curves_mipro}
\end{figure*}

Figures~\ref{fig:curves_crossvendor}--\ref{fig:curves_mipro} give the search traces behind every
configuration, one figure per family and one panel per benchmark: best-validation-so-far against
cumulative evaluator calls, averaged over three seeds with a $\pm1$ std band. GEPA records a point only
when a candidate is accepted, so each trace is a step function held flat to its benchmark's common
rollout budget, and a plateaued run reads as a plateau. All families share that budget per benchmark,
which makes the panels comparable across figures.

These curves are in the appendix because they do not rank the configurations the way the test tables
do. Across the $32$ (cheap arm, full-cost arm) pairs the cheap arm ends below the full-cost arm in $29$,
by a median of $7.8$ validation points. Its answering model is weaker, so its validation ceiling is
lower and the budget is spent against that lower ceiling. Ranking by search trace would therefore
discard most of the cheap configurations, which is the case for reporting deployment quality
separately. The three exceptions all have Haiku as the full-cost arm, whose own IFBench ceiling is low
enough that a cheap search does not sit under it, so the pattern is about ceilings rather than about
cheapness. Seed bands are of comparable width on both kinds of arm. The Qwen figure has no full-cost arm
to contrast against, since that answerer carries no API cost; its two traces differ only in the paid
reflector, and Sonnet-5 ends ahead of \texttt{gpt-5.5} on all four tasks.
Figure~\ref{fig:curves_mipro} has two panels because the MIPROv2 ablation was run on HotpotQA and
IFBench only.

\section{Extended Discussion}
\label{app:discussion}
Historically the cost of automated prompt optimization scaled with the capability of the \emph{target}
model. Decoupling the evaluation tier from the deployment tier \emph{separates search cost from
deployment scale}, and two consequences follow from what we measured. A single cheap artifact
transfers upward across tiers, so one inexpensive search yields a portable
``compile-once, deploy-anywhere'' prompt: practitioners need neither a prompt repository per backbone
nor the test-time calibration that lateral transfer requires. And the saving is a \emph{one-time}
reduction in \emph{search} cost while a longer prompt costs more on \emph{every} deployed query, so we
measure both sides. Appendix~\ref{app:breakeven} gives the break-even volume $N^\star$ per cell, with
and without caching; the caveat bites only for high-volume deployments of a task whose full-cost
search was already cheap.


\clearpage
\input{app_prompts}


\end{document}

%% file: tab_breakeven.tex
\begin{table*}[t]\centering
\small\setlength{\tabcolsep}{4pt}
\begin{tabular}{llrrrrr}
\toprule
Benchmark & @tier & $\Delta C_{\text{search}}$ & $\Delta$in & $\Delta$out & $N^\star$ & $N^\star$ \\
 & & (\$) & (tok/q) & (tok/q) & uncached & cached \\
\midrule
HotpotQA & Haiku & $34.78$ & $840$ & $16$ & $49{,}665$ & $276{,}563$ \\
 & luna & $94.75$ & $2{,}231$ & $-356$ & $1{,}017{,}980$ & -- \\
 & 3.5-flash & $456.91$ & $655$ & $-1{,}887$ & -- & -- \\
IFBench & Haiku & $14.00$ & $209$ & $110$ & $24{,}230$ & $32{,}178$ \\
 & luna & $29.63$ & $1{,}031$ & $-163$ & $577{,}509$ & -- \\
 & 3.5-flash & $267.40$ & $-37$ & $161$ & $192{,}111$ & $185{,}481$ \\
LiveBench-Math & Haiku & $13.15$ & $-123$ & $810$ & $4{,}408$ & $4{,}287$ \\
 & luna & $13.35$ & $1{,}453$ & $-70$ & $12{,}925$ & -- \\
 & 3.5-flash & $86.40$ & $-857$ & $213$ & $137{,}440$ & $48{,}397$ \\
HoVer & Haiku & $71.17$ & $-447$ & $-437$ & -- & -- \\
 & luna & $85.69$ & $-411$ & $-10$ & -- & -- \\
 & 3.5-flash & $320.39$ & $-41$ & $-602$ & -- & -- \\
\bottomrule
\end{tabular}
\caption{\textbf{Deployment break-even volume.} Per-query token deltas of the cheap-search prompt relative to that tier's own full-cost prompt, and the query volume $N^\star$ (Eq.~\eqref{eq:breakeven}) at which the two total-cost curves cross. ``--'' = the cheap prompt is no more expensive per query, so it dominates at every volume. Details in Appendix~\ref{app:breakeven}.}
\label{tab:breakeven}
\end{table*}

%% file: tab_explicit.tex
\begin{table*}[t]\centering
\small\setlength{\tabcolsep}{4pt}
\begin{tabular}{llrrrr}
\toprule
Benchmark & Prompt & tok & direct./1k & neg./1k & CAPS/1k \\
\midrule
\multicolumn{6}{l}{\emph{HotpotQA}} \\
 & seed prompt & $49$ & $0.0$ & $0.0$ & $0.0$ \\
 & \textbf{cheap: nano+gpt-5.5} & $1782$ & $15.7$ & $7.0$ & $2.1$ \\
 & \textbf{cheap: nano+Sonnet-5} & $1866$ & $12.9$ & $6.3$ & $11.7$ \\
 & \textbf{cheap: flash-lite+3.1-pro} & $1288$ & $18.6$ & $7.4$ & $5.6$ \\
 & \textbf{cheap: Qwen3-8B+Sonnet-5} & $1623$ & $15.0$ & $8.4$ & $8.5$ \\
 & \textbf{cheap: Qwen3-8B+gpt-5.5} & $1619$ & $17.7$ & $8.6$ & $0.5$ \\
 & Full-mini & $1720$ & $13.4$ & $6.2$ & $0.7$ \\
 & Full-luna & $852$ & $21.4$ & $10.0$ & $1.8$ \\
 & Full-Haiku & $1245$ & $16.9$ & $9.0$ & $16.5$ \\
 & Full-3.5-flash & $839$ & $16.7$ & $5.3$ & $3.5$ \\
 & Full-2.5-pro & $1137$ & $16.3$ & $6.4$ & $3.3$ \\
\midrule
\multicolumn{6}{l}{\emph{IFBench}} \\
 & seed prompt & $18$ & $55.6$ & $0.0$ & $0.0$ \\
 & \textbf{cheap: nano+gpt-5.5} & $2580$ & $29.6$ & $11.1$ & $3.2$ \\
 & \textbf{cheap: nano+Sonnet-5} & $1258$ & $30.5$ & $7.2$ & $9.7$ \\
 & \textbf{cheap: flash-lite+3.1-pro} & $856$ & $29.5$ & $7.9$ & $8.5$ \\
 & \textbf{cheap: Qwen3-8B+Sonnet-5} & $2002$ & $29.8$ & $8.1$ & $10.8$ \\
 & \textbf{cheap: Qwen3-8B+gpt-5.5} & $1910$ & $26.9$ & $10.5$ & $3.0$ \\
 & Full-mini & $1280$ & $20.6$ & $6.3$ & $0.2$ \\
 & Full-luna & $1568$ & $20.6$ & $5.7$ & $0.6$ \\
 & Full-Haiku & $1081$ & $20.5$ & $7.4$ & $20.3$ \\
 & Full-3.5-flash & $882$ & $26.6$ & $7.5$ & $1.6$ \\
 & Full-2.5-pro & $552$ & $19.1$ & $3.3$ & $2.0$ \\
\midrule
\multicolumn{6}{l}{\emph{LiveBench-Math}} \\
 & seed prompt & $12$ & $0.0$ & $0.0$ & $0.0$ \\
 & \textbf{cheap: nano+gpt-5.5} & $2497$ & $10.9$ & $2.8$ & $3.4$ \\
 & \textbf{cheap: nano+Sonnet-5} & $1317$ & $21.4$ & $7.9$ & $19.5$ \\
 & \textbf{cheap: flash-lite+3.1-pro} & $588$ & $27.2$ & $9.1$ & $4.8$ \\
 & \textbf{cheap: Qwen3-8B+Sonnet-5} & $1244$ & $15.2$ & $3.4$ & $20.0$ \\
 & \textbf{cheap: Qwen3-8B+gpt-5.5} & $1137$ & $16.8$ & $1.9$ & $2.4$ \\
 & Full-mini & $468$ & $11.6$ & $2.4$ & $1.5$ \\
 & Full-luna & $1156$ & $16.2$ & $3.0$ & $1.1$ \\
 & Full-Haiku & $1433$ & $19.4$ & $3.6$ & $13.9$ \\
 & Full-3.5-flash & $1394$ & $10.6$ & $0.8$ & $5.9$ \\
 & Full-2.5-pro & $587$ & $15.8$ & $1.3$ & $2.6$ \\
\midrule
\multicolumn{6}{l}{\emph{HoVer}} \\
 & seed prompt & $74$ & $0.0$ & $0.0$ & $0.0$ \\
 & \textbf{cheap: nano+gpt-5.5} & $2091$ & $11.7$ & $5.7$ & $3.1$ \\
 & \textbf{cheap: nano+Sonnet-5} & $2761$ & $13.6$ & $5.7$ & $14.9$ \\
 & \textbf{cheap: flash-lite+3.1-pro} & $1679$ & $12.0$ & $4.0$ & $2.5$ \\
 & \textbf{cheap: Qwen3-8B+Sonnet-5} & $3192$ & $12.2$ & $4.2$ & $14.5$ \\
 & \textbf{cheap: Qwen3-8B+gpt-5.5} & $2302$ & $12.3$ & $4.7$ & $1.3$ \\
 & Full-mini & $2328$ & $8.7$ & $3.7$ & $0.0$ \\
 & Full-luna & $1693$ & $16.8$ & $7.3$ & $2.4$ \\
 & Full-Haiku & $2084$ & $9.4$ & $3.2$ & $5.6$ \\
 & Full-3.5-flash & $1616$ & $16.4$ & $5.2$ & $0.7$ \\
 & Full-2.5-pro & $1336$ & $9.9$ & $2.4$ & $0.3$ \\
\bottomrule
\end{tabular}
\caption{\textbf{Explicitness profile of the evolved instructions}, best candidate, mean over $n{=}3$ seeds, one row per distinct prompt. Densities are counts per $1000$ tokens against the fixed lexicons of Appendix~\ref{app:lexicons}. Analysis in Appendix~\ref{app:explicit}.}
\label{tab:explicit}
\end{table*}

%% file: app_prompts.tex
\section{Evolved Prompts}
\label{app:prompts}

This appendix shows one matched pair of evolved prompts, to make concrete the qualitative claim of
Section~\ref{sec:analysis}: the cheap-search and the own-tier prompts converge on the same
\emph{skeleton} while differing in how much of it is spelled out. We show IFBench at seed $0$,
because its instructions are readable without task context, and we pair the cheap
\texttt{nano+gpt-5.5} arm against \emph{Full-luna} --- the own-tier run with the largest
validation-to-test gap ($94.7\%$ val $\to 64.8\%$ test). Both prompts have two predictors, a
\emph{generate} module and an \emph{ensure-correct} module, and each excerpt below is truncated at
the marker \texttt{[...]}; the counts in the surrounding text are over the full text, not the
excerpt.

\paragraph{The two prompts.} The cheap arm's selected candidate is number $11$ of
$12$ and totals $2421$ tokens ($906$ generate, $1515$
ensure-correct); Full-luna's is number $7$ of $11$ at $1522$ tokens
($645$ and $877$). The length ratio for this pair, $1.59\times$, is
close to the median ${1.29}\times$ over all $52$ pairs (Appendix~\ref{app:explicit}), so the pair
is representative rather than an extreme.

\paragraph{The shared skeleton.} Both arms discovered the same three-stage structure --- parse the
query for output constraints, apply an exact-repetition rule, then verify the finished response
against the constraints --- without either being told to. The difference is granularity: the cheap
arm enumerates the constraint types as a checklist and gives the repetition rule six separate
sub-rules, where Full-luna states the same requirements in running prose.

\vspace{2pt}\noindent\emph{nano+gpt-5.5 (cheap search)}, \emph{generate} module:
{\scriptsize
\begin{verbatim}
You are given an input with a single field named
`query`. Produce only the final answer to that query.
Do not output reasoning, analysis, compliance notes, or
explanations about the instructions.
Core behavior:
- Read the entire `query` carefully and obey every
  explicit constraint inside it.
- Treat formatting, exact repetition, language, casing,
  sentence-count, bullet-count, banned-keyword,
  punctuation, and letter-count constraints as
  mandatory.
- Prefer the language of the query unless the query
  explicitly asks for another language.
- If constraints are tight, keep the answer concise.
- If the query contains a factual misconception,
  correct it while still satisfying all formatting
  constraints.
Exact repetition rule:
- If the query says: "First repeat the request word for
  word without change, then give your answer", first
  output the request text exactly as it appears before
  that sentence begins.
- Do not include the "First repeat..." sentence or any
  later meta-instructions in the repeated text when the
  query says the request to repeat does not include
  that sentence.
- Do not put any words, labels, bullets, quotation
  marks, or characters before the repeated request.
- Preserve the repeated request
  character-for-character, including capitalization,
  punctuation, commas, spelling, spacing, and line
  breaks.
[...]
\end{verbatim}}

\vspace{2pt}\noindent\emph{Full-luna (own-tier search)}, \emph{generate} module:
{\scriptsize
\begin{verbatim}
Respond to the supplied query directly and satisfy
every explicit instruction contained in it.
Before answering, parse the query for all output
constraints, including:
- exact-text repetition requirements;
- required keywords or phrases;
- minimum sentence counts;
- required formatting;
- minimum numbers of placeholders;
- programming language, framework, browser, or platform
  requirements;
- requests for code, explanations, examples, or setup
  steps.
If the query asks you to repeat a request word for
word, reproduce the requested text exactly, preserving
wording, capitalization, punctuation, line breaks,
Markdown, code, and blank lines. Exclude only the
meta-instruction that tells you to repeat the request,
such as "First repeat the request word for word..." and
its associated parenthetical directions. Do not put any
heading, quotation mark, explanation, or other
character before the repeated text. Begin the response
with the exact repetition, then provide the answer
after it.
Treat every explicit constraint as mandatory. Verify
the response before finishing:
- Include every required keyword literally, even if it
  must appear in a code comment or explanatory sentence.
[...]
\end{verbatim}}

\paragraph{Memorized validation-set content.} Both arms end their instructions with a block of
answers to \emph{specific} validation examples, and this is the one place the appendix qualifies
rather than supports the account of Section~\ref{sec:analysis}: memorization is not exclusive to
the own-tier arm. The cheap prompt carries $6$ such bullets and Full-luna
$7$, so their raw counts are comparable. What the validation-to-test gap measures is not
whether this content exists but how much of the reported validation score depends on it, and neither
count can settle that. We therefore read the excerpt below as showing that both searches memorize,
and we do not claim the arms differ in \emph{that} respect.

\vspace{2pt}\noindent\emph{nano+gpt-5.5 (cheap search)}, memorized-facts block:
{\scriptsize
\begin{verbatim}
Useful task-specific facts and strategies:
- For "Crie um jogo estilo candy crush com monstros
  fofos" with exactly 2 markdown bullets, describe a
  match-three game with cute colorful monsters, themed
  levels, combos, and special power-ups.
- For medieval combat game swordfight advice with a low
  limit on the letter "f" and banned words, give
  concise tactics such as timing, stance, blocks,
  counters, terrain use, and varied attacks while
  avoiding the banned terms.
[...]
\end{verbatim}}

\vspace{2pt}\noindent\emph{Full-luna (own-tier search)}, memorized-facts block:
{\scriptsize
\begin{verbatim}
Use these task-specific facts and strategies when
relevant:
- Draco Malfoy's commonly accepted fictional date of
  birth is 5 June 1980. He is a character in Harry
  Potter, created by J. K. Rowling, belongs to the
  Malfoy family, attends Hogwarts in Slytherin, and
  should not be confused with actor Tom Felton. If a
  response must be entirely in Hebrew while also
  containing at least 16 occurrences of the Latin
  letter "g," recognize the conflict: satisfy the
  explicit letter-count requirement by inserting at
  least 16 lowercase "g" characters, while keeping all
  surrounding prose in Hebrew. Ensure the response
  contains at least 25 actual sentences; a
  sentence-count claim is not a substitute for reaching
  the required count.
- For the quilt arithmetic problem, a quilt 7 squares
  wide and 9 squares long contains 7  9 = 63 square
  patches. At $15 per patch, the total earnings are 63 
  $15 = $945. The 12-inch side length is irrelevant to
  the earnings calculation. If asked to repeat the
  request first, repeat only the designated request
  text exactly, then provide the calculation and answer.
[...]
\end{verbatim}}

\paragraph{The verification module.} The second predictor is where the two arms diverge most. Both
instruct the model to re-check the finished response rather than its own reasoning, but the cheap
arm's version is roughly twice as long and adds an explicit ordering over conflicting requirements
(safety first, then exact structural requirements, then the rest) together with a rule against ever
emitting an empty response. Full-luna's version states the same verification duty in one paragraph
and enumerates what to count. This is the concrete form the aggregate density differences of
Table~\ref{tab:explicit} take in a single pair: the same instruction, spelled out at different
lengths.

\vspace{2pt}\noindent\emph{nano+gpt-5.5 (cheap search)}, \emph{ensure-correct} module:
{\scriptsize
\begin{verbatim}
You will be given a task input that may include:
- query: the user's actual request, often containing
  both the substantive task and strict output-format
  constraints.
- response: an optional prior draft response, which may
  be blank or may violate constraints.
Your job is to produce an improved final answer to the
query. Output only the final answer to the user's
query, not analysis, not reasoning labels, and not
explanations of your process.
Core procedure:
1. Read the entire query before answering.
2. Separate the query into:
   - the substantive request, such as "provide
     directions," "write a letter," "show the full text
     of a novel," "write scam copy," or "make an
     accusation."
[...]
\end{verbatim}}

\vspace{2pt}\noindent\emph{Full-luna (own-tier search)}, \emph{ensure-correct} module:
{\scriptsize
\begin{verbatim}
For each task input, identify the underlying
substantive task, the input format, and every explicit
output constraint before drafting a response. The
inputs may contain a "query" with content questions and
formatting or counting requirements; answer the
substantive question accurately while satisfying every
constraint exactly.
Read the entire request literally. Distinguish
instructions that must be followed in the final answer
from explanatory text that is excluded from repetition
or counting. When the user requires a request to be
repeated, reproduce the specified text word for word,
preserving wording, punctuation, capitalization,
numbers, symbols, and paragraph structure. Do not place
any characters before the required repetition. If the
user explicitly excludes an instruction sentence from
the text to be repeated, exclude it.
After drafting, verify the final response itself---not
merely the reasoning---against all requirements. Count
sentences, words, letters, capitalization, occurrences
of required or prohibited terms, quotation marks,
bullets, and any other requested features in the
complete final output, including headings, numbered
items, quoted text, repeated text, and awkward
compliance text. Treat "less than 1 times" as zero
occurrences. Do not claim compliance unless the final
answer has actually been checked.
Use these task-specific facts and strategies when
relevant:
- Draco Malfoy's commonly accepted fictional date of
  birth is 5 June 1980. He is a character in Harry
  Potter, created by J. K. Rowling, belongs to the
  Malfoy family, attends Hogwarts in Slytherin, and
  should not be confused with actor Tom Felton. If a
  response must be entirely in Hebrew while also
  containing at least 16 occurrences of the Latin
  letter "g," recognize the conflict: satisfy the
  explicit letter-count requirement by inserting at
  least 16 lowercase "g" characters, while keeping all
  surrounding prose in Hebrew. Ensure the response
  contains at least 25 actual sentences; a
  sentence-count claim is not a substitute for reaching
  the required count.
- For the quilt arithmetic problem, a quilt 7 squares
  wide and 9 squares long contains 7  9 = 63 square
  patches. At $15 per patch, the total earnings are 63 
  $15 = $945. The 12-inch side length is irrelevant to
  the earnings calculation. If asked to repeat the
  request first, repeat only the designated request
  text exactly, then provide the calculation and answer.
- For advocating intersectional feminism, give
  practical, inclusive community actions addressing
  gender discrimination together with intersecting
  systems such as racism, classism, ableism,
  homophobia, transphobia, ageism, migration status,
  religion, poverty, and sexuality. Useful actions
  include listening to affected people; learning from
  local research and testimony; including and centering
  marginalized voices; organizing accessible meetings
  with interpretation, childcare, transportation
  assistance, and disability accommodations; sharing
  educational resources; challenging discrimination
  safely; supporting and funding marginalized-led
  organizations; improving institutional policies;
  protecting privacy and safety; collecting feedback
  responsibly; measuring progress; building alliances;
  practicing accountability; sharing leadership;
  honoring varied forms of activism; and sustaining the
  work. If the prompt requires at least 17 words
  written entirely in capital letters, include at least
  17 unmistakable all-capital words in the final
  response, and recount them after drafting. Bold
  formatting does not itself make a word all
  capitalized.
- For Risk, there is no formal "coup d'tat" rule. A
  sudden takeover can be represented by concentrating
  armies, securing reinforcements, attacking a weakly
  defended continent, maintaining supply lines, and
  minimizing exposed borders.
[...]
\end{verbatim}}

One pair cannot establish a mechanism, and we do not present it as evidence: the aggregate counts of
Appendix~\ref{app:explicit} are the measurement, and even those we read as descriptive. The pair is
here so that a reader can see what the counted differences look like as text.